\documentclass[twocolumn,5p]{elsarticle}

\usepackage{epsfig}
\usepackage{graphicx}
\usepackage{amsmath}
\usepackage{bm}
\usepackage{amssymb}
\usepackage{footnote}
\usepackage{times}
\usepackage{helvet}

\usepackage{courier}
\usepackage{color}
\usepackage{natbib}

\usepackage{multirow}
\usepackage[algoruled,vlined,linesnumbered]{algorithm2e}
\usepackage{algorithmic} 

\newcommand{\zhu}{\textcolor[rgb]{0,0,0}}

\journal{Journal of Artificial Intelligence}

\begin{document}

\begin{frontmatter}

\title{
	Discovering Visual Concept Structure with Sparse and Incomplete Tags
	}

\author[mymainaddress]{Jingya Wang}
\ead{jingya.wang@qmul.ac.uk}

\author[mymainaddress]{Xiatian Zhu}\ead{xiatian.zhu@qmul.ac.uk}
\author[mymainaddress]{Shaogang Gong}\ead{s.gong@qmul.ac.uk}

\address[mymainaddress]
{School of EECS, 
	Queen Mary University of London, Mile End Road, London, E1 4NS, UK}

\begin{abstract}
Discovering automatically the semantic structure of tagged visual data (e.g. web videos and images) is important for visual data analysis and interpretation, enabling the machine intelligence for effectively processing the fast-growing amount of multi-media data. However, this is non-trivial due to the need for jointly learning underlying correlations between heterogeneous visual and tag data. The task is made more challenging by inherently sparse and incomplete tags. In this work, we develop a method for modelling the inherent visual data concept structures based on a novel Hierarchical-Multi-Label Random Forest model capable of correlating structured visual and tag information so as to more accurately interpret the visual semantics, e.g. disclosing meaningful visual groups with similar high-level concepts, and recovering missing tags for individual visual data samples. Specifically, our model exploits hierarchically structured tags of different semantic abstractness and multiple tag statistical correlations in addition to modelling visual and tag interactions. As a result, our model is able to discover more accurate semantic correlation between textual tags and visual features, and finally providing favourable visual semantics interpretation even with highly sparse and incomplete tags. We demonstrate the advantages of our proposed approach in two fundamental applications, visual data clustering and missing tag completion, on benchmarking video (i.e. TRECVID MED 2011) and image (i.e. NUS-WIDE) datasets.

\end{abstract}


\end{frontmatter}


\section{Introduction}
\label{sec:intro}

\begin{figure*} [t!]
	\centering
	\includegraphics[width=1\linewidth]{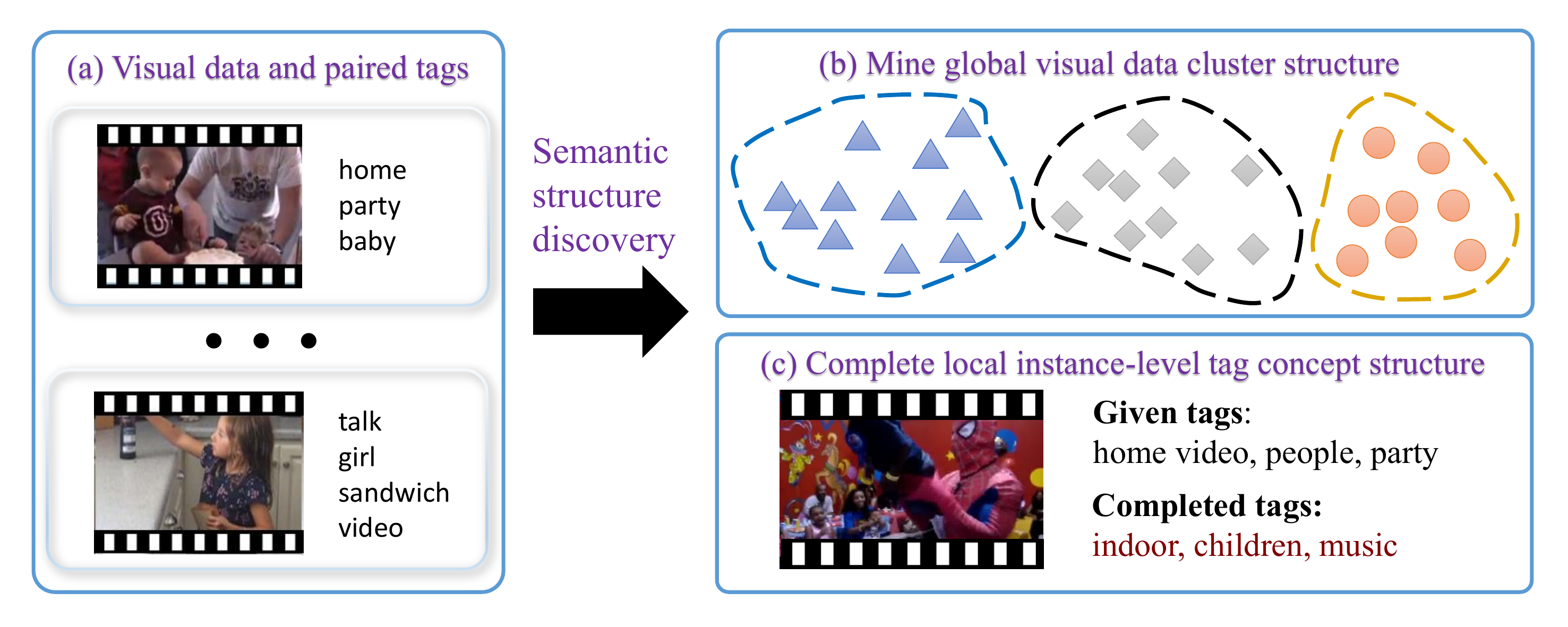}
	\vskip -1cm
	\zhu{
		\caption{
			Problem illustration: we aim to develop an automated visual semantics discovery approach by exploiting (a) both visual and sparse tag data for 
			(b) mining global visual data cluster structure and 
			(c) completing local instance-level tag concept structure.
		}
	}
	\label{fig:problem}
\end{figure*}

A critical task in visual 
data analysis is to automatically discover and interpret
the underlying semantic concept structure
of large quantities of data effectively and quickly,
which allows the computing intelligence for automated organisation and management of large scale multi-media data.
However, semantic structure discovery for visual data by visual feature analysis alone is
inherently limited due to the semantic gap 
between low-level visual features and high-level semantics, 
particularly under the ``curse'' of high dimensionality,
{\color{black} where visual features are often represented in a
  high-dimensional feature space} \cite{beyer1999nearest}.
On the other hand, videos and images are often attached with additional non-visual data,
e.g. typically some textual sketch (Figure \ref{fig:problem}(a)). 
Such text information can include short tags contributed by either
users or content providers, for instance,  
videos/images from the YouTube and Flickr websites. 
\zhu{Often, tags may provide uncontrolled mixed levels of information
	but being also incomplete with respect to the visual content.
	This motivates (1) {\em multi-modality based data cluster discovery}  
	(where visual data samples in each hidden cluster/group share the same underlying high-level concept relevant to both visual appearance and textural tags in a latent unknown space) \cite{jain2010data,zhou2013latent,vahdat2014discovering},
	and (2) {\em instance-level tag structure completion} 
	(where the tag set is defined as the combination of all presented tags
	and
	missing tag revelation for each visual data sample may rely on both visual appearance and given tags) \cite{wu2013tag,feng2014image,lin2013image}.
	The former considers global data group structure, 
	{\color{black} e.g. data clustering (Figure \ref{fig:problem}(b)) that
		serves as a critical automated data analysis strategy 
		with important fundamental applications, such as 
		summarising video data for automatically removing redundancy and discovering
		meaningful / interesting content patterns
		hidden in large scale data corpus without any human labelling effort \cite{truong2007video},
		detecting anomalies and salient data \cite{jain2010data},
		or facilitating unstructured data browsing and examination \cite{vahdat2014discovering}.}
	In contrast, the latter
	addresses local tag label structure of 
	individual visual instances,
	e.g. tag completion (Figure \ref{fig:problem}(c)) that aims to
	automatically recover missing concepts presented in visual data.
	In this multi-modality data learning context,
	it is necessary to highlight and distinguish three fundamental notions:
	(1) visual content, (2) visual features, and (3) textual tags.
	Among them, the latter two are different representations of the former, i.e. visual content --
	the actual target data/objects of our problem.
	By visual concept structure, we particularly refer to 
	the concept structure of ``visual content''
	rather than ``visual features''.}

Exploiting readily accessible textual tags in visual
content interpretation has shown to be beneficial \cite{zhou2013latent,vahdat2014discovering,feng2014image}.
Nonetheless, existing methods are restricted in a number of ways:
(1) Tags are assumed with similar abstractness
(or flattened tag structure). Intrinsic hierarchical tag
structures are ignored in model design;
(2) Tag statistical correlations and interactions between
visual and tag data are not fully exploited, partly due to
model complexity and design limitation. 
Incorporating such information into existing models effectively is not straightforward.

In general, joint learning of visual and text information, two different
heterogeneous data modalities, in a shared representational space
is non-trivial because:
(1) The heteroscedasticity problem~\cite{Duin_PAMI04}, that is,
disparate data modalities significantly differ in representation (continuous or categorical)
and distribution characteristics with different scales and covariances.
In addition, the dimensionality of visual data often exceeds that of
tag data by a large extent, like thousands vs. tens/hundreds. 
Because of this dimensionality discrepancy problem, 
a simple concatenation of heterogeneous feature spaces may result in a
incoherent representation favourably inclined towards one dominant modality
data and leading to suboptimal results.
(2) Visual features can be inaccurate and unreliable, 
	due to the inherently ambiguous and noisy visual data, 
	and the imperfect nature of feature extraction. 
	It is challenging to suppress the negative influence of unknown noisy visual features in data structure modelling.
(3) The available text tags are often sparse and
incomplete.  
This causes an inevitable problem that 
the visual (with much richer but also noisier and redundant information) 
and tag (being often sparse and incomplete although complementary) data are not always 
completely aligned and correlated.

In this work, we develop a model for robust visual semantic structure discovery and interpretation by
employing both visual features and available sparse/incomplete
text tags associated with the videos/images.
The {\bf contributions} of this work are as follows:
{\bf (I)} We formulate a novel approach
capable of effectively extracting and fusing information from 
ambiguous/noisy visual features and sparse/incomplete textual tags
for precisely discovering and mining the inherent visual semantic structures.
This is made possible by introducing a new Hierarchical-Multi-Label
Random Forest (HML-RF) model with a reformulated information gain
function that allows to model the interactions between visual features 
and incomplete tags simultaneously.
Specifically, our model is designed to minimise the uncertainty of tag distributions 
in an ``abstract-to-specific'' hierarchical fashion so as to 
exploit the high-order skeletal guidance knowledge embedded in tag hierarchy structure.
{\bf (II)} We introduce a unified 
tag dependency based algorithm 
to cope with the tag sparseness and incompleteness problem.
In particular, we formulate a principled way of 
locally integrating multiple statistical correlations (co-occurrence and mutual-exclusion) 
among tags during model optimisation. 
{\bf (III)} We develop a data clustering method based on the proposed HML-RF model
by measuring pairwise similarity between visual samples for accurately discovering
the semantic global group structure of all visual data.
{\bf (IV)} We design three HML-RF tree structure driven tag prediction algorithms to recover missing tags for completing the local tag concept structure of individual visual data instances.
We demonstrated the efficacy and superiority of our proposed approach 
on the TRECVID MED 2011 \cite{vahdat2014discovering} (web videos)
and NUS-WIDE \cite{nus-wide-civr09} (web images) datasets 
through extensive comparisons with 
related state-of-the-art clustering, multi-view learning and tag completion methods.

\section{Related Work}
\label{sec:rela}

We review contemporary related studies on global structure analysis (e.g. data clustering) and local concept structure recovery (e.g. missing tag completion) using tagged visual data, tag correlation and hierarchy, and random forest models. 

\vspace{0.1cm}
\noindent {\bf Tagged visual data structure analysis}:
Compared with low-level visual features, textual information provides
high-level semantic meanings which can help bridge the gap between video
features and human cognition.  
Textual tags have been widely employed along with visual features
to help solve a variety of challenging computer vision problems, such as
visual recognition \cite{vahdat2013handling} and retrieval \cite{natarajan2012multimodal}, image annotation \cite{makadia2008new}.
Rather than these supervised methods,
we focus on 
\zhu{structurally-constrained learning approach without the
need of particular human labelling}.
Whilst a simple combination of visual features and textural tags may give
rise to the difficult heteroscedasticity problem, 
Huang et al. \cite{huang2012affinity} alternatively
seek an optimal combination of similarity measures derived \textcolor{black}{from} different data modalities. 
The fused pairwise similarity can be then utilised for data clustering by existing graph based clustering algorithms such as spectral clustering \cite{ng2002spectral}.
As the interaction between visual appearance and textual tags is not modelled in
the raw feature space but on the similarity graphs, the information loss
in graph construction can not be recovered.   
Also, this model considers no inter-tag correlation.  

\zhu{
Alternatively, multi-view learning/embedding methods
are also able to jointly learn visual and text data by inferring a latent common subspace, 
such as 
multi-view metric learning \citep{quadrianto2011learning},
Restricted Boltzmann Machine 
and auto-encoders \citep{srivastava2012multimodal,ngiam2011multimodal},
visual-semantic embedding \cite{frome2013devise},
Canonical Correlation Analysis (CCA) and its variants \cite{hardoon2004canonical,hwang2012learning,gong2014multi,rai2009multi,sharma2012generalized}.
Inspired by the huge success of deep neural networks, 
recently a few works have attempted to combine deep feature learning and CCA for 
advancing multi-view/modality data modelling \cite{andrew2013deep,wang2015deep}.
However, these methods usually assume a reasonably large number of tags available. 
Otherwise, the learned subspace may be subject to sub-optimal cross-modal correlation,
e.g. in the case of significantly sparse tags.
In addition, whilst incomplete tags can be considered as a special case of noisy labels, 
	existing noise-tolerant methods   \cite{natarajan2013learning,sukhbaatar2014training,frenay2014classification}
	are not directly applicable. 
	This is because they usually handle classification problems where
	a separate training dataset is required for model building,
	which however is not available in our context.}

More recently, Zhou et al. \cite{zhou2013latent} 
devised a Latent Maximum Margin Clustering (Latent MMC) model 
for assisting tagged video grouping. 
This model separates the whole task into two isolated stages: 
tag model learning and clustering, and thus their interaction is ignored.
To tackle the above problem, 
Arash et al. \cite{vahdat2014discovering} proposed a Structural MMC model 
where the correlations between visual features, tags and clusters are jointly modelled and optimised.
The best results of clustering tagged videos are attained by 
Flip MMC \cite{vahdat2014discovering} with the idea of flipping tags mainly for
addressing the tag sparseness problem.
In both MMC variants, tags are organised and used in a flat structure, 
whilst different tags may correspond to varying degrees of concept abstractness.
Further, the statistical correlations between tags are neglected during optimisation.
These factors may cause either degraded data modelling or knowledge
loss, as shown in our experiments.
Compared with these existing methods above, the proposed approach in this
work is capable of jointly considering
interactions between visual and tag data modalities, 
tag abstractness hierarchical
structure and tag statistical correlations 
within a unified single model.

\vspace{0.1cm}
\noindent \textbf{Missing tag completion}:
Text tags associated with videos and images 
are often sparse and incomplete, particularly those provided by web users.
This may impose negative influence on tag-based applications and thus
requires effective methods for tag completion.
Different from conventional tag annotation \cite{cabral2011matrix,mu2011accelerated},
tag completion does not require an extra completely annotated training dataset.
Liu et al. \cite{liu2012image}
formulated tag completion as a non-negative data factorisation problem.
Their method decomposes the global image representation into regional tag representations, on which the appearance of individual tags is characterised and visual-tag consistency is enforced.
Wu et al. \cite{wu2013tag}
performed tag recovery by searching for the optimal tag matrix which maximises the consistency with partially observed tags, visual similarity (e.g. visually similar samples are constrained to have common tags) and tag co-occurrence correlation.
Lin et al. \cite{lin2013image} 
developed a sparsity based tag matrix reconstruction method 
jointly considering visual-visual similarity, visual-tag association and tag-tag concurrence in completion optimisation.
Similarly, Feng et al. \cite{feng2014image} 
proposed another tag matrix recovery approach based on the low rank matrix theory \cite{candes2009exact}. 
Visual-tag consistency is also integrated into optimisation by exploring the graph Laplacian technique.
However, all these methods ignore tag abstractness hierarchy structure, which
may affect negatively the tag correlation and visual consistency
modelling. 
Additionally, they depend on either global or regional visual
similarity measures which can suffer from unknown noisy visual
features or incomplete tags. 
Compared with these existing methods, 
we investigate an alternative strategy for tag completion,
that is, to discover visual concept structure for identifying meaningful
neighbourhoods and more accurate tag inference.  
To that end, we formulate a new Hierarchical-Multi-Label
Random Forest (HML-RF) capable of jointly modelling tag and visual
data, exploiting the intrinsic tag hierarchy knowledge, and the
inherent strengths of a random forest for feature selection. 
We compare quantitatively our method with the state-of-the-art
alternative tag completion models in extensive experiments and
demonstrate the clear advantages of the proposed HML-RF model (Section
\ref{sec:evaluate_completion}).

\vspace{0.1cm}
\noindent \textbf{Tag hierarchy and correlations}:
Hierarchy (a pyramid structure) is a natural knowledge
organisation structure of our physical world, from more abstract to
more specific in a top-down order
\cite{fellbaum1998wordnet,deng2009imagenet},
and has been widely used in numerous studies, 
for example tag recommendation \cite{shepitsen2008personalized},
semantic image segmentation \cite{zheng2014dense},
and object recognition \cite{deng2014large}.
%
Typically, an accurate hierarchy structure is assumed and utilised \cite{zheng2014dense,deng2014large}.
But this is not always available, e.g.
tag data extracted from some loosely structured meta-data source can only
provide a rough hierarchy with potentially inaccurate relations,
as the meta-data associated with videos in the TRECVID dataset.
So are the user-provided tags from social media websites like Flickr.
Such noisy hierarchy imposes more challenges but still useful if used properly.
To that end, we exploit hierarchical tag structures in a more
robust and coherent way for effective semantic structure modelling 
of sparsely tagged video/image data. 


One of the most useful information encoded in hierarchy 
is inter-tag correlation, and \textit{co-occurrence}
should be most widely exploited,
e.g. 
image annotation~\cite{griffiths2005infinite,chen2010efficient},
and object classification~\cite{deng2014large}.
This positive label relation is useful since it provides 
a context for structuring the complexity of the real-world concepts/things.
In contrast, \textit{mutual-exclusion} is another (although less popular) relation between concepts. 
As opposite to co-occurrence, it is negative but complementary.
Its application includes object detection~\cite{choi2010exploiting,desai2011discriminative}, 
multi-label image annotation~\cite{chen2011multi}, 
multi-task learning~\cite{zhou2010exclusive},
and object recognition~\cite{deng2014large}.
Unlike the above supervised settings, \zhu{we investigate both correlations in
a {\em structurally-constrained learning} manner.}
Also, we do not assume their availability as in the case
of~\cite{deng2014large}. Instead, we automatically mine these
correlations from sparsely labelled data.
Different from~\cite{chen2011multi} where the tag structure is regarded as flat, 
we consider the co-occurrence and mutual-exclusive correlation between tags across layers of the tag hierarchy. 
We learn this pairwise
relation, rather than assuming as prior knowledge as in~\cite{deng2014large}.
Further, we relax the stringent assumption of accurate tags as made in
\cite{choi2010exploiting,desai2011discriminative,chen2011multi}
and the model is designed specifically to tolerate tag incompleteness and sparseness.
	Our goal is to exploit automatically the tag correlations and
        the available tag hierarchy structure effectively for inferring
        semantics on visual data and discovering visual concept structures.

\vspace{0.1cm}
\noindent \textbf{Random forest models}:
Random forests have been shown to be effective for many computer vision tasks
\cite{Criminisi2012,zhu2016constrained,zhu2013constrained,zhu2014constructing}. 
Below we review several most related random forest variants.
Montillo et al. \cite{montillo2011entangled} presented an Entangled Decision Forest for helping
image segmentation by 
propagating knowledge across layers, 
e.g. dependencies between pixels and objects.  
Recently, Zhao et al. \cite{zhao2014unified} proposed a multi-task forest for face analysis
via learning different tasks at distinct layers according to the correlations between multi-tasks 
(e.g. head pose, facial landmarks).
All these models are supervised. 
\zhu{In contrast, our forest model performs structurally-constrained learning since we aim to
	discover and obtain semantic data structure using heterogeneous tags
	that are not target category labels but merely some semantic constraints.}
Furthermore, our model is unique in its capability of handling missing data, which 
is not considered in \cite{zhao2014unified,montillo2011entangled}.
The Constrained Clustering Forest (CC-Forest) \cite{zhu2013video,zhu2016learning}
is the most related to our HML-RF model, in that it is also utilised for data structure analysis e.g. measuring data affinity.
The advantage of our model over CC-Forest are two-folds:
(1) The capability for exploiting the tag hierarchical structure knowledge
and (2) The superior effectiveness of tackling missing data, as shown in our experiments (Section \ref{sec:exp}).


\section{Methodology}
\label{sec:mod}
\noindent {\bf Rational for model design}: 
We want to formulate a unified visual semantic structure discovery 
model capable of 
addressing the aforementioned challenges and limitations of existing methods.
Specifically, to mitigate the heteroscedasticity and dimension discrepancy problems,
we need to isolate different characteristics of visual and tag data, 
yet can still fully exploit the individual modalities 
as well as cross-modality interactions in a balanced manner.
For handling tag sparseness and incompleteness, 
we propose to utilise the constraint information derived from inter-tag statistical correlations
\cite{griffiths2005infinite,choi2010exploiting,deng2014large}.
To that end, we wish to explore random forest~\cite{BreimanML01,shi2006unsupervised,Criminisi2012} 
because of:
(1) Its flexible training objective function for facilitating multi-modal data modelling and reformulation;
(2) The decision tree's hierarchical structures for flexible integration of abstract-to-specific structured tag topology;
(3) Its inherent feature selection mechanism for handling inevitable data noise.
Also, we need to resolve several shortcomings of the conventional clustering forest 
\cite{shi2006unsupervised}, as in its original form it is not best suited for solving our
problems in an unsupervised way. 
Specifically, 
clustering forest expects a fully concatenated representation 
as input during model training, it therefore
does not allow a balanced utilisation of two modalities simultaneously 
(the dimension discrepancy problem), 
nor exploit interactions between visual and tag
features. 
The existing classification forest is also not suitable
as it is supervised and aims to learn a 
prediction function with class labelled training data (usually a
single type of tag)~\cite{BreimanML01}. Typical video/image tags do not
offer class category labels.
However, it is interesting to us that in contrast to the clustering forest, 
the classification forest offers a more balanced structure for 
using visual (as split variables) and tag (as semantic evaluation)
data that is required for tackling the heteroscedasticity problem by isolating
the two heterogeneous modalities during learning.

\begin{figure*} [th!]
	\centering
	\includegraphics[width=1\linewidth]{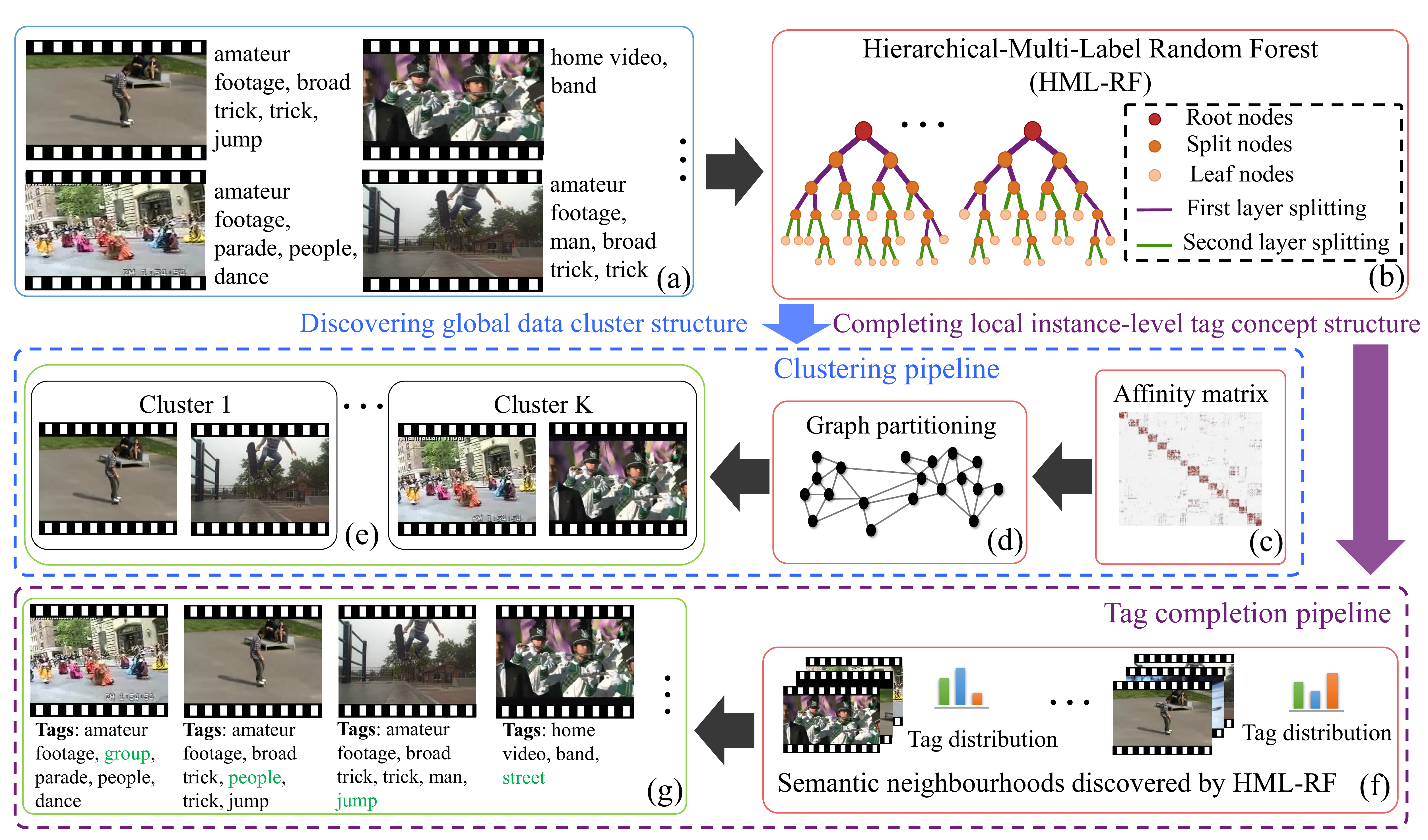}
	\vskip -.7cm
	\textcolor{black}{
		\caption{
			An overview of the proposed visual semantic structure discovery approach.
			(a) Example videos and associated tags; 
			(b) The proposed HML-RF model designed to exploit inherent tag hierarchy for modelling correlations between ambiguous visual features and sparse tags, discover visual concept structures in two aspects:
			\textbf{Discovering global data cluster structure:}
			(c) Semantically constrained affinity matrix induced by HML-RF $\rightarrow$
			(d) Graph-based clustering to discover semantic groups $\rightarrow$ 
			(e) Resulting clusters with semantic similarity despite significant
			visual disparity.
			\textbf{Completing local instance-level tag concept structure: }
			(f) Semantic neighbourhood structures discovered by the proposed HML-RF model, 
			which can then be exploited for (g) inferring missing tags to complete local concept structure at the data sample level.
		}
	}
	\label{fig:pipeline}
\end{figure*}

\vspace{0.1cm}
\noindent {\bf Approach overview}: 
We want to reformulate the classification forest for
automatically disclosing the semantic structure of videos or images with tags.
To that end, we propose a novel \textit{Hierarchical-Multi-Label Random Forest} (HML-RF).
Our model goes beyond the classification
forest in the following aspects: 
(1) Employing tags to constrain tree structure learning, rather than learning a generalised prediction function as~\cite{BreimanML01,Criminisi2012};
(2) Introducing a new objective function allowing \textit{acceptance of multi-tags}, 
\textit{exploitation of abstract-to-specific tag hierarchy} and
\textit{accommodation of multiple tag correlations} simultaneously. 
Instead of learning a classifier, 
HML-RF is designed to infer visual semantic concept structure
for more accurately revealing both global visual data group structures and 
local tag structures of individual visual data samples.
These structural relationships among data samples imply their underlying data group/cluster relations (obtained using a standard graph based clustering algorithm on the similarity graph 
estimated by our HML-RF model),
as well as the specific tag concept structures of individual samples
(predicted using the discovered semantic neighbourhoods encoded in the tree structures of HML-RF).
An overview of the proposed visual concept structure discovery approach is depicted in Figure \ref{fig:pipeline}.

\vspace{0.1cm}
\noindent {\bf Notations}: 
We consider two data modalities,
(1) \textit{Visual data modality} - 
We extract a $\mathit{d}$-dimensional visual descriptor from the $i$-th video/image sample denoted by $ \bm{x}_i = \left( x_{i,1}, \dots, x_{i,\mathit{d}} \right) \in {R}^{\mathit{d}}, i = 1, \dots, $ {$n $}. All visual features are formed as $X = \{\bm{x}_i\}_{i=1}^n$.
(2) \textit{Tag data modality} -  
Tags associated with videos/images are extracted from the meta-data files or given by independent users. 
We represent $\mathit{m}$ types of binary tag data ($Z = \{1,\dots,m\}$) attached 
with the $i$-th video/image as $\bm{y}_i = \left( y_{i,1},
\dots, y_{i,\mathit{m}} \right) \in [0,1]^\mathit{m}$. 
All tag data is defined as $Y = \{\bm{y}_i\}_{i=1}^n$.
More details are provided in Section \ref{sec:dataset_settings}.

\subsection{Conventional Random Forests}
\label{sec:RF}
Let us briefly introduce conventional random forests before detailing the proposed HML-RF model.

\noindent \textbf{Classification forests}:
A classification forest~\cite{BreimanML01} 
contains an ensemble of 
{$\tau$}
	binary decision trees.
Growing a decision tree involves a recursive node splitting procedure 
until some stopping criterion is satisfied.
The training of each split node is a process of binary split function optimisation, defined as
\begin{equation}
	\mathrm{h}(\bm{x}, \bm{w}) = 
    \left\{
	\begin{array}{l l}
    	0 	& \quad \text{if ${x}_{f} < \theta$,} \\
		1 	& \quad \text{otherwise.} \\
	\end{array} \right.
\label{eqn:split_function}
\end{equation}
with two parameters 
{$\bm{w} = \left[f, \theta\right]$}: 
(i) a feature dimension {${x}_{f}$} with {$f \in \{1,\dots, d\}$}, and 
(ii) a feature threshold 
{$\theta$}. 
The optimal split parameter $\bm{w}^*$ is chosen via
\begin{equation}
\bm{w}^* = \text{argmax}_{\bm{w} \in W} \Delta \psi_\text{sl},
\label{eqn:split_parameter_optimisation}
\end{equation}
where 
\zhu{
the parameter search space 
$W = {\left\{ \bm{w}_i \right\}_{i=1}^{\nu_\mathrm{try}(|S|-1)}}$ is formed 
by enumerating the threshold (or cut-point) on each of 
$\nu_\mathrm{try}$ randomly selected features (without replacement),
with $S$ denoting the sample set reaching the split node $s$.
More specifically, the cut-points of each feature are defined as
the unique midpoints of the intervals between ordered values from this feature on samples $S$.
Thus, there is $|S|-1$ candidate cut-points for every chosen feature, 
with $|\cdot|$ referring to the cardinality of a set.}
The information gain 
$\Delta \psi_\text{sl}$  is formulated as
\begin{equation}
\Delta \psi_\text{sl} = \psi_s - 
\frac{|L|}{|S|} \psi_l - \frac{|R|}{|S|} \psi_r,
\label{eqn:info_gain}
\end{equation}
where $L$ and $R$ denote the data set routed into the left $l$ and right $r$ children, 
and $L \cup R = S$. 
The uncertainty { $\psi$ } over the label distribution
can be computed as 
the Gini impurity~\cite{breiman1984classification} or entropy \cite{Criminisi2012}.
We used the former in our HML-RF model due to its simplicity and efficiency,
i.e. the complexity of computing $\psi_\text{sl}$ is $O(1)$ as it is 
computed over the label distribution.

\vspace{0.1cm}
\noindent \textbf{Clustering forests}:  
Clustering forests aim to obtain an optimal data partitioning
based on which pairwise similarity measures between samples can be inferred.
In contrast to classification forests, 
clustering forests require no ground truth label information during the training phase.
Similarly, a clustering forest consists of binary decision trees. 
The leaf nodes in each tree define a spatial partitioning of the training data.
Interestingly, the training of a clustering forest can be performed
using the classification forest optimisation approach 
by adopting the pseudo two-class algorithm~\cite{BreimanML01,shi2006unsupervised}.
With this data augmentation strategy, 
the clustering problem becomes a canonical classification problem 
that can be solved by the classification forest training method as discussed above. 
The key idea behind this algorithm is 
to partition the augmented data space into dense and sparse regions 
\cite{LiuCIKM2000}.
One limitation of clustering forests is the limited ability in mining multiple modalities,
as shown in Section \ref{sec:exp}.

\subsection{Hierarchical-Multi-Label Random Forest}
\label{sec:HMLRF}

Our HML-RF can be considered as an extended hybrid model of classification and clustering forests.
The model inputs include visual features $\bm{x}$ and tag data $\bm{y}$ of visual data samples 
(analogous to classification forest), 
and the output is semantic tree structures 
which can be used to predict an affinity matrix $\bm{A}$ over input samples $X$ (similar to clustering forest).
Conventional classification forests \cite{BreimanML01} typically 
assume single label type. 
In contrast, HML-RF can accept multiple types simultaneously as follows.
%

\vspace{0.1cm}
\noindent \textbf{Accommodating multiple tags}:
A HML-RF model uses visual features as splitting variables 
to grow trees (HML-trees) as in Equation~(\ref{eqn:split_function}), 
but exploits all types of tag data {\em together} as tree structuring constraints
in optimising
{
$\bm{w} = \left[f, \theta\right]$. 
}
Formally, we extend the conventional single-label based information
gain function Equation (\ref{eqn:info_gain}) to  multi-labels for training HML-trees:
\begin{align}
\label{eqn:our_infogain_multitag}
	\Delta \psi_{\text{ml}} = 
	\sum_{i = 1}^{\mathit{m}} \Delta \psi_\text{sl}^{i} 
\end{align}
This summation merges all individual information gains $\Delta \psi_\text{sl}^i$
from the $i$-th tag in an intuitive way for simultaneously enforcing knowledge 
of multiple tags into the HML-tree training process.
Hence, the split functions are optimised in a similar way as supervised classification forests, 
and semantics from multiple tags are enforced simultaneously.

\vspace{.1cm}
\noindent {\em \underline{Discussion}}: 
In the context of structure discovery, e.g. tagged video/image clustering,
\zhu{ 
it should be noted that our way of exploiting tags is different from conventional 
supervised classification forests 
since the tags are not target classes but semantic 
constraints.}
We call this ``\textit{structurally-constrained learning}''.
Additionally, the interactions between visual features (on which split functions are defined) 
and tags (used to optimise split functions) are also modelled during learning 
by identifying the most discriminative visual features
w.r.t. a collection of textual tags.
Importantly, this separation of visual and tag data solves naturally
the dimensionality discrepancy problem and addresses the heteroscedasticity challenge.
Moreover, HML-RF benefits from the feature selection mechanism inherent to random forest for coping with noisy visual data by selecting the most discriminative localised split
functions (Equation (\ref{eqn:split_function})) over multiple tags
simultaneously.	
%

\vspace{0.1cm}
\noindent \textbf{Incorporating tag hierarchy}:
Equation~(\ref{eqn:our_infogain_multitag}) implies that all the tags have
similar abstractness, as all of them are used in every
split node (i.e. a flatten structure of tags). 
\zhu{However, diverse tags may 
	lie in multiple abstractness layers and how to exploit this information
	is critical for visual data structure modelling.
	The intuition is that
	tag hierarchy encodes approximately some relation knowledge between different 
	underlying data structures and likely provides useful high-order skeletal guidance 
	during the data structure inference process.
	The tag hierarchy structure can be roughly available from data source or 
	automatically estimated by text analysis(see Section \ref{sec:dataset_settings}).
To further exploit the abstractness guidance information in tag hierarchy,
we introduce an adaptive
hierarchical multi-label information gain function as:}
\begin{equation}
\label{eqn:our_infogain_hierarchy}
	\Delta \psi_{\text{hml}} = 
	\sum_{k=1}^{\mu}  \left(  \prod_{j=1}^{k-1} (1 - \alpha_j)  \alpha_k \sum_{i \in Z_k} \Delta \psi_\text{sl}^{i} \right) 
\end{equation}
where 
{$Z_k$} denotes the tag index set of the 
{$k$-th} layer in the tag hierarchy (totally 
{$\mu$} layers), 
with 
$\cup_{k=1}^{\mu} Z_k = Z$, and $\forall_{j \neq k} Z_j \cap Z_k = \emptyset$.
Binary flag 
{$\alpha_k \in \{0, 1\}$} indicates the impurity of the 
{$k$-th} tag layer, 
{$k \in \{1,\dots,\mu\}$}, i.e.
{$\alpha_k = 0$} when tag values are identical, i.e. pure,
across all the training samples $S$ of split node $s$
in any tag $i \in Z_k$, 
{$\alpha_k = 1$} otherwise.
{\color{black} Note, $\alpha$ is designed to be non-continuous 
so HML-tree per-node optimisation can focus on mining the underlying 
interactive information of visual-textual data
at one specific semantic abstractness level.
This shares a similar spirit to the ``divide-and-conquer'' learning strategy,
e.g. reducing the local learning difficulty by considering {\em first}
more homogeneous concepts only in training individual weak tree node
models, {\em before} finally making the whole model to capture better
semantic structure information. This is in contrast to solving the
more difficult holistic optimisation problem on the entire tag
set with a mixture of different abstractness levels.} 
The target layer is {$k$} in case that 
{$\alpha_k = 1$} and 
{$\forall \alpha_j = 0, 0 < j < k$}.

\vspace{.1cm}
\noindent {\em \underline{Discussion}}:
\zhu{This layer-wise design allows the data partition optimisation 
to concentrate on the \textit{most abstract} and \textit{impure}
tag layer (i.e. the target layer) 
so that the abstractness skeletal information in the
tag hierarchy can be 
gradually embedded into the top-down HML-tree growing procedure
for guiding the interaction modelling between visual and tag data in an abstract-to-specific fashion. 
This design and integration shall be natural and coherent 
because both tag hierarchy and HML-tree model are in the shape of pyramid
and the divide-and-conquer modelling behaviour in HML-RF is intuitively suitable for the abstract-to-specific tag structure.}
We will show the empirical effectiveness of this layer-wise information gain design in our experiments (Section \ref{sec:exp_model_analysis}). 

\vspace{0.1cm}
\noindent \textbf{Handling tag sparseness and incompleteness}: 
We further improve the HML-RF model by employing tag statistical correlations for addressing
tag sparseness problem, as follows:
We wish to utilise the dependences among tags to infer
missing tags with a confidence measure (continuous soft tags), 
and exploit them along with labelled (binary hard) tags 
in localised split node optimisation, e.g. 
Equations (\ref{eqn:info_gain}) and (\ref{eqn:our_infogain_hierarchy}).

In particular, two tag correlations are considered: 
\textit{co-occurrence} - often co-occur in the same video/image samples thus positively correlated,  and
\textit{mutual-exclusion} - rarely simultaneously appear 
so negatively correlated.
They are complementary to each other, since for a particular sample, 
co-occurrence helps predict the \textit{presence} degree of some missing tag
based on another frequently co-occurrent tag who is labelled,
whilst mutual-exclusion can estimate the \textit{absence} degree of a tag according to its negative relation with another labelled tag.
Therefore, we infer tag positive $\{\hat{y}^+_{.,i}\}$ and 
negative $\{\hat{y}^-_{.,i}\}$ confidence scores
based upon tag co-occurrent and mutual-exclusive correlations, respectively.
Note that  $\{\hat{y}^+_{.,i}\}$ and $\{\hat{y}^-_{.,i}\}$ 
are not necessarily binary
but more likely real number, e.g. $[0, 1]$.
In our layered optimisation, 
we restrict the notion of missing tag to samples
$S_\textrm{miss} = \{\mathring{\bm{x}}\}$ where
no tag in the target layer is labelled, 
and consider cross-layer tag correlations considering that
a hierarchy is typically shaped as a pyramid, 
with more specific 
tag categories at lower layers where likely more labelled tags are available.
Suppose we compute the correlations between the tag 
{$i \in Z_k$} (the target tag layer) and the tag 
{$j \in \{Z_{k+1},\dots,Z_{\mu}\}$
} (subordinate tag layers).

\noindent \textit{Co-occurrence}:
We compute the co-occurrence {$\varrho_{i,j}$} as 
\begin{equation}
\label{eqn:cooccur}
	\varrho_{i,j} = {co_{i,j}}/{o_j},
\end{equation}
where $co_{i,j}$ denotes the co-occurrence frequency of tags $i$ and
$j$, that is, occurrences when both tags simultaneously appear in the
same video/image across all samples; 
and $o_j$ denotes the number of occurrences of tag $j$ over all samples.
Note that these statistics are collected from the available tags.
The denominator 
{${o_j}$} here is used to down-weight over-popular tags $j$:
Those often appear across the dataset, 
and their existence thus gives a weak positive cue of 
supporting the simultaneous presence of tag $i$. 
For example, tag `people' may appear in most videos 
and so brings a limited positive correlation to others.
In spirit, this design shares the principle of Term Frequency Inverse Document
Frequency \cite{berger2000bridging,Sivic03}, 
which considers the inverse influence of total term occurrence times 
across the entire dataset as well.
Once 
{$\varrho_{i,j}$} is obtained, for a potentially missing tag 
{$i \in Z_k$} of {$\mathring{\bm{x}}$} $\in S_\textrm{miss}$, we estimate its positive score 
{$\hat{y}^+_{\cdot,i}$} via: 
\begin{equation}
\label{eqn:positive_score}
	\hat{y}^+_{\cdot,i} = \sum_{j \in \{Z_{k+1},\dots,Z_{\mu}\}} \varrho_{i,j} y_{\cdot,j}
\end{equation}
where $y_{\cdot,j}$ refers to the $j$-th tag value of $\mathring{\bm{x}}$.
With Equation~(\ref{eqn:positive_score}), we accumulate the positive support from all labelled subordinate tags to estimate the presence confidence of tag $i$.

\begin{algorithm} [h] \footnotesize
	\caption{\footnotesize Split function optimisation in a HML-tree}
	\label{Alg:learn_HMLRF}	 
	\SetAlgoLined
	\KwIn
	{ At a split node $s$ of a HML-tree $t$: \\
		\quad - Visual data $X_s$ of training samples $S$ arriving at $s$; \\ 
		\quad - Corresponding labelled tag data $Y_s$; \\
		\quad - Soft tag estimation using tag correlations: \\
		\quad \quad * Positive scores $\{\hat{y}^+_{.,i}\}$ 
		estimated with Equations~(\ref{eqn:cooccur}) and (\ref{eqn:positive_score});\\
		\quad \quad * Negative scores $\{\hat{y}^-_{.,i}\}$
		estimated with Equations~(\ref{eqn:excl}) and (\ref{eqn:negative_score});\\
	}
	\KwOut
	{\\
		\quad - The best feature cut-point $\bm{w}^*$; \\
		\quad - The associated child node partition $\{ L^*, R^* \}$; \\
	}
	
	\textbf{Optimisation}: \\
	Initialise $L^* = R^* = \emptyset$, $\Delta \psi_{\text{hml}}^*=0$, $\bm{w}^*=[-1,-\infty]$; \\
	\For{$k \leftarrow 1$ \KwTo $\nu_\mathrm{try}$}
	{
		Select a visual feature $x_k \in \{ 1,\dots,d \}$ randomly \zhu{without replacement}; \\
		\For {each possible cut-point of $x_k$}
		{
			Split $S$ into a candidate partition $\{ L, R \}$; \\
			Compute $\Delta \psi_{\text{hml}}$ with Equations~(\ref{eqn:info_gain}) and~(\ref{eqn:our_infogain_hierarchy}); \\
			\If{$\Delta \psi_{\text{hml}} > \Delta \psi_{\text{hml}}^*$}
			{
				Update $\bm{w}^*$ with $x_k$ and current threshold; \\
				Update $\Delta \psi_{\text{hml}}^* = \Delta \psi_{\text{hml}}$, $L^* = L$, and $R^* = R$. \\
			}
		}
	}
\end{algorithm}

\vspace{0.1cm}
\noindent \textit{Mutual-exclusion}:
We calculate this negative correlation as 
\begin{equation}
\label{eqn:excl}
	\epsilon_{i,j} = \text{max}(0, r^{-+}_{i,j} - r^-_i) / (1 - r^-_i),
\end{equation}
where
$r^-_i$ refers to the negative sample percentage on tag $i$ across all samples,
and $r^{-+}_{i,j}$ the negative sample percentage on tag $i$ over samples with positive tag $j$.
The denominator $(1 - r^-_i)$ is the normalisation factor.
Hence, $\epsilon_{i,j}$ measures statistically the relative increase in 
negative sample percentage on tag $i$ 
given positive tag $j$. 
This definition reflects statistical exclusive degree of tag $j$ against tag $i$ intuitively.
The cases of $\epsilon < 0$ are not considered since they are already measured in the co-occurrence.  
Similarly, we predict the negative score 
$\hat{y}^-_{\cdot,i}$ for $\mathring{\bm{x}}$ on tag $i$ with:
\begin{equation}
\label{eqn:negative_score}
	\hat{y}^-_{\cdot,i} = \sum_{j \in \{Z_{k+1},\dots,Z_{\mu}\}} \epsilon_{i,j} y_{\cdot,j}.
\end{equation}

%

Finally, we normalise both 
{$\hat{y}^+_{\cdot,i}$} and 
{$\hat{y}^-_{\cdot,i}$, ${i \in Z_p}$, } into the unit range $[0, 1]$.
%
%
Algorithm~\ref{Alg:learn_HMLRF}
summarises the split function optimisation procedure in a HML-tree.

\subsection{Discovering Global Data Cluster Structure}
\label{sec:clustering}

Our HML-RF model is designed to discover visual semantic structures, 
e.g. global group structure over data samples. 
Inspired by 
clustering forests~\cite{BreimanML01,shi2006unsupervised,Criminisi2012},
this can be achieved by first estimating pairwise proximity between samples 
and then applying graph based clustering methods to obtain data groups (Figure \ref{fig:pipeline}(c,d,e)).

\vspace{0.1cm}
\noindent \textbf{Inducing affinity graph from the trained HML-RF model}: 
Specifically,
the $t$-th ($t \in \{1,\dots,\tau\}$) 
tree within the HML-RF model 
partitions the training samples at its leaves. 
Each leaf node forms a {\em neighbourhood}, which contains 
a subset of data samples that share visual and semantic commonalities.
All samples in a neighbourhood are {\em neighbours} to each other.
These neighbours are considered similar both visually and semantically 
due to the proposed split function design (Equation (\ref{eqn:our_infogain_hierarchy})). 
More importantly,
tag correlations and tag hierarchy structure knowledge are also taken into account
in quantifying the semantic concept relationships.
With these neighbourhoods, 
we consider an affinity model without any parameter to tune.
Specifically, we assign pairwise similarity 
``$1$'' 
for sample pair ($\bm{x}_i$, $\bm{x}_j$) 
if they fall into the same HML-tree leaf node (i.e. being neighbours), 
and 
``$0$'' 
otherwise. 
This results in a tree-level affinity matrix $\bm{A}^t$.
A smooth affinity matrix $\bm{A}$ can be obtained through
averaging all the tree-level affinity matrices:
\begin{equation}
\bm{A} = \frac{1}{\tau}\sum_{t=1}^{\tau} \bm{A}^t
\label{eqn:ensembel}
\end{equation}
with $\tau$ the tree number of HML-RF.
Equation~(\ref{eqn:ensembel}) is adopted as the ensemble model of HML-RF due to
its advantage of suppressing the noisy tree predictions, although other alternatives as the product of tree-level predictions are possible~\cite{Criminisi2012}.
Intuitively, the multi-modality learning strategies of HML-RF enable
its data similarity measure to be more meaningful.
This can benefit significantly video/image clustering using a graph-based 
clustering method, as described next.
\vspace{0.1cm}
\noindent \textbf{Forming global clusters}:
Once the affinity matrix $\bm{A}$ is obtained, one can apply any off-the-shelf graph-based clustering model to 
acquire the final clustering result, e.g. spectral clustering~\cite{ng2002spectral}.
Specifically, we firstly construct a sparse $\kappa$-NN graph,
(Figure \ref{fig:pipeline}(d)), 
whose edge weights are defined by {$\bm{A}$} (Figure~\ref{fig:pipeline}(c)).
Subsequently, we symmetrically normalise $\bm{A}$ to obtain
$\bm{S} = \bm{D}^{-\frac{1}{2}} \bm{A} \bm{D}^{-\frac{1}{2}}$, 
where $ \bm{D} $ denotes a diagonal degree matrix with elements 
$\bm{D}_{i,i} = \sum_{j=1}^{n}{\bm{A}_{i,j}}$ ($n$ denotes the video/image sample number).
Given $\bm{S}$, we perform spectral clustering to 
discover the latent clusters of videos/images (Figure \ref{fig:pipeline}(e)).
Each sample $\bm{x}_i$ is then 
assigned to 
a cluster index $c_i \in {C}$, 
where ${C} = \{1,\dots,p\}$ contains a total of $p$ cluster indices.

\subsection{Completing Local Instance-Level Concept Structure}
\label{sec:method_tag_completion}

In addition to inferring the global group structure, the learned
semantic structure by the HML-RF model can also be exploited for reasoning the local concept structures of individual samples which are often partial and incomplete due to sparsely labelled tags. 
This task is known as {\em tag completion} \cite{wu2013tag}.
Intuitively, the potential benefit of HML-RF for tag completion is due to semantic neighbourhoods over data samples formed during the model training phase (Section \ref{sec:HMLRF}).
More specifically, as data splits in HML-RF consider both correlations between visual features and tags, and dependencies between tags in abstractness hierarchy and statistics,
visually similar neighbour samples (e.g. sharing the same leaves) 
may enjoy common semantic context and/or tags, 
and thus helpful and indicative in recovering missing tags. Formally, we aim to predict the existence probability $p(\bm{x}_*, j)$
of a missing tag $j \in Z$ in a sample $\bm{x}_*$.  
Given estimated $p(\bm{x}_*, j)$, those with top probabilities are considered as missing tags.
To that end, we derive three tree-structure driven missing tag completion algorithms as below. 

\vspace{0.1cm}
\noindent
{\bf (I) Completion by local neighbourhoods}: 
We estimate $p(\bm{x}_*, j)$ by local neighbourhoods formed in HML-RF.
Specifically, 
we first 
identify the neighbourhood $N^t$ of $\bm{x}_*$ in each HML-tree $t \in \{1,2,\dots,\tau\}$ by retrieving the leaf node that $\bm{x}_*$ falls into.
Second, for each $N^t_{\bm{x}_*}$, 
we compute the distribution $\text{pdf}(t, j)$ of tag $j$ over $\bm{x}_*$'s neighbours. 
As these neighbours are similar to $\bm{x}_*$, we use $\text{pdf}(t,
j)$ as a tree-level prediction.
However, some neighbourhoods are unreliable due to the inherent visual
ambiguity and tag sparseness, we thus ignore them and consider only
confident ones with 
$\text{pdf}(t, j) = 0$ (called negative neighbourhood) or 
$\text{pdf}(t, j) = 1$ (called positive neighbourhood).
Finally, we calculate $p(\bm{x}_*, j)$ as
\begin{equation}
p(\bm{x}_*, j) = \frac{|P_{j}^{+}|}{|P_{j}^{+}| + |P_{j}^{-}|}
\label{eqn:completion}
\end{equation}
where $|P_{j}^{+}|$ and $|P_{j}^{-}|$ are the sets of
positive and negative neighbourhoods, respectively.
As such, the negative impact of unreliable neighbourhoods 
can be well suppressed.
We denote this Local Neighbourhoods based method as ``{\bf HML-RF(LN)}''.

\vspace{0.1cm}
\noindent 
{\bf (II) Completion by global structure}:
Similar to local neighbourhoods of HML-RF,
the data clusters (obtained with the method as described in Section \ref{sec:clustering}) can be considered as global neighbourhoods.
Therefore, we may alternatively exploit them for missing tag prediction.
In particular, we assume that $\bm{x}_*$ is assigned with cluster $c$. 
We utilise the cluster-level data distribution for missing tag estimation as:
\begin{equation}
p(\bm{x}_*, j) = \frac{|X_c^{+}|}{|X_c| - 1}
\label{eqn:clust_pred}
\end{equation}
where $X_c$ are data samples in cluster $c$,
and $X_c^{+} \subset X_c$ are samples with labelled positive tag $j$.
The intuition is that visual samples from the same cluster (thus of same high-level semantics/concept) are likely to share similar tags. 
Note, this is also a tree-structure based inference method in that
these clusters are induced from tree-structure driven 
similarity measures (Section \ref{sec:clustering}).
We denote this Global Cluster based prediction algorithm as ``{\bf HML-RF(GC)}''.

\vspace{0.1cm}
\noindent
	{\bf (III) Completion by affinity measure}:
	Similar to k-nearest neighbour classification \cite{cover1967nearest,weinberger2009distance},
	we perform tag completion using affinity measures. 
	Specifically, we utilise the tag information of $\kappa$ nearest neighbours $N_{\kappa}$ by adaptive weighting:
	\begin{equation}
	p(\bm{x}_*, j) = \frac{1}{|{\kappa}|} 
	\sum_{i \in N_{\kappa}} y_{i,j} \bm{A_{i,*}}
	\label{eqn:tag_cpl_AM}
	\end{equation}
	where $y_{i,j}$ denotes the tag $j$ value of the $i$-th nearest neighbour $\bm{x}_i$,
	$\bm{A_{i,*}}$ is the pairwise similarity between $\bm{x}_i$
        and $\bm{x}_*$ estimated by Equation (\ref{eqn:ensembel}), or
        the weight. 
	Different from HML-RF(LN) that models the individual neighbourhoods within tree leaves, this method considers weighted pairwise relationship across all HML-trees, i.e. how many times two samples fall into the same leaf nodes.
	Conceptually, this can be considered as a hybrid model of
        HML-RF(LN) and HML-RF(GC) due to the inherent relation with
        both local neighbourhoods (i.e. tree leaves) and global
        clusters (the same similarity estimation).  
	We denote this HML-RF Affinity Measure based tag recovery algorithm as ``{\bf HML-RF(AM)}''.

\section{Experiments}
\label{sec:exp}

\subsection{Datasets and Experimental Settings}
\label{sec:dataset_settings}
\begin{figure*} 
	\centering
	\includegraphics[width=0.75\linewidth]{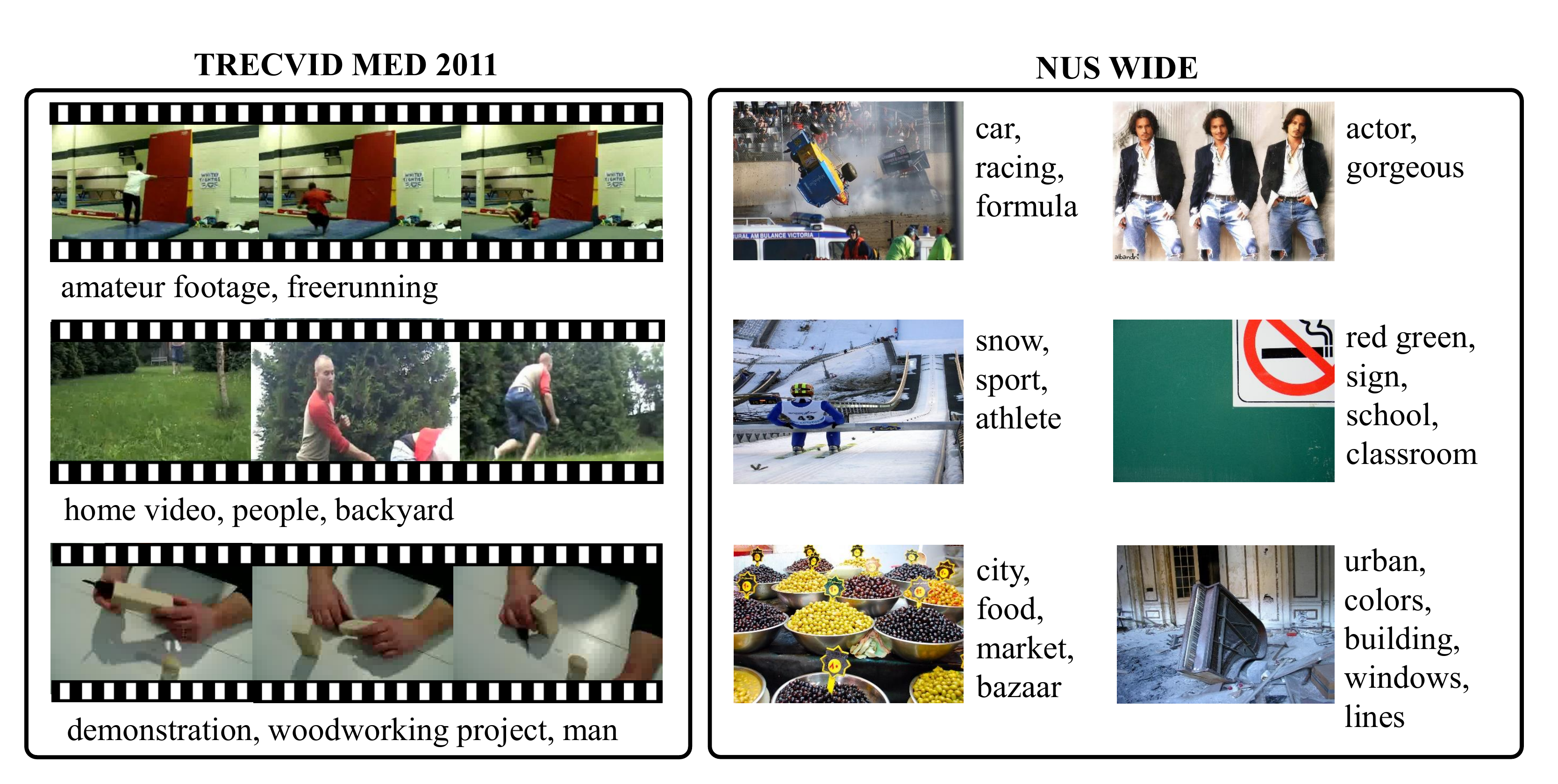}
	\vspace{-0.5cm}
	\caption{
		Examples from TRECVID MED 2011 \cite{over2011trecvid} and NUS-WIDE \cite{nus-wide-civr09}.
	 } 
	\label{fig:dataset}
\end{figure*}

\noindent \textbf{Datasets}:
We utilised two web-data benchmarks, 
the TRECVID MED $2011$ video dataset \cite{over2011trecvid} and 
the NUS-WIDE image dataset \cite{nus-wide-civr09},
for evaluating the performance of our proposed HML-RF model.
Figure \ref{fig:dataset} shows a number of samples from the two datasets.

\vspace{0.1cm}
\noindent {\em TRECVID MED 2011}:
It contains $2379$ web videos from $15$ clusters 
which we aim to discover in global structure analysis
as in \cite{zhou2013latent,vahdat2014discovering}.
This dataset is challenging for clustering using only visual features, in that 
videos with the same high-level concepts can present significant variety/dynamics in visual appearance.
This necessitates the assistance of other data modalities, 
e.g. tags automatically extracted from textual judgement files associated with video samples \cite{vahdat2014discovering}.
Specifically, a total of $114$ tags were obtained and used in our evaluation.
On average, around $4$ tags ($3.5\%$ of all tags) were extracted per video, 
thus very sparse and incomplete with the need for recovering many unknown missing tags.
\zhu{
The tag hierarchy was established according to the structure presented in the meta-data files with two levels of tag abstractness. For example, tag ``party'' is more structurally abstract than tags ``people/food/park'' in the context of TRECVID videos where a number of semantic events (e.g.  with respect to wedding ceremony and birthday celebration) may be meaningfully related with tag ``party'' whilst tags ``people/food/park'' should be very general and common to many different events and thus structurally specific.}
{\color{black} For video clustering, we aim to discover the underlying
  event category groups of web videos, given the ground-truth
  annotation available. This is similar to that of 
  \cite{zhao2008efficient,vahdat2014discovering}.}   
For evaluating the performance of missing tag completion,
we manually completed a subset of $200$ video samples on $51$ 
randomly selected tags as ground truth \cite{feng2014image}.

\vspace{0.1cm}
\noindent {\em NUS-WIDE}:
\zhu{We further evaluated the HML-RF model on 
a tagged web image dataset, NUS-WIDE \cite{nus-wide-civr09}.
We randomly sampled $30$ clusters, each of which contains over 500 images and
a total of $17523$ images were selected for the
evaluation of both global image clustering and local
tag concept completion.
This dataset contains $1000$ different tags.
Every image is labelled with $4.8$ tags (i.e. $0.48\%$ of all tags) on average.
}

\zhu{
For NUS-WIDE, we need to establish the tag hierarchy since 
tags are given in a flat structure. 
Inspired by \cite{wei2006lda,gong2014multi}, we estimate the tag abstractness degree by
mining and employing tag-image data statistics information. 
To be more precise, we first apply
term frequency inverse document frequency (tf-idf) weighting
to the binary tag vector $\bm{y}_i = [y_{i,1},\dots,y_{i,m}]$ 
of each image $i$
($m$ denotes the tag type number),
and get a new tag representation 
$\tilde{\bm{y}}_i = [\tilde{y}_{i,1},\dots,\tilde{y}_{i,m}]$.
This allows $\tilde{\bm{y}}_i$ %
encoding the importance of each tag
against the corresponding image by taking
into account the tag-image statistic relation among the entire dataset. 
Then, we perform K-means over these tf-idf weighted tag vectors $\{\tilde{\bm{y}}_i\}$ of all images
to obtain ${E}$ topic clusters.
In each cluster $e$ where $\{\tilde{\bm{y}}_i^e\}$ fall into, 
we compute the abstractness or representativeness score for tag $j$
as $\sigma_j^e = \sum \tilde{{y}}_{i,j}^e$
and select the tags 
with top-$\eta$ highest $\sigma_j^e$ scores
into the current hierarchy layer.
By performing this selection on all clusters,
we form the current layer with selected most abstract tags 
whilst the remaining tags drop into lower layers.
Similarly, we build one or multiple lower hierarchy layers 
on the remaining tags with the same steps above.
Actually, we can consider this tag hierarchy formation 
as a process of revealing underlying topics
in a layer-wise fashion.
We select more tags per cluster for lower layers considering 
the potentially pyramid hierarchy shape,
e.g. choosing top $\eta = 3\times i$ tags from every cluster for the $i$-th hierarchy layer.
On tagged NUS-WIDE images,  
tag ``race'' is considered more structurally abstract than 
tags ``sky/street/house/men''
by our proposed method above.
This is reasonable because 
there exist some underlying groups (e.g. regarding Formula-1 and raft competition)
that are semantically relevant with tag ``race''
whilst tags ``sky/street/house/men'' describe concrete objects 
that may be possibly shared by much more different data structures 
and hence structurally specific.
Our proposed HML-RF model is formulated particularly
to accommodate such abstractness skeletal knowledge
in rough tag hierarchy
for discovering and interpreting 
sparsely and/or incompletely tagged visual data,
beyond conventional multi-modality correlation learning methods
that often attempt to straightly correlate visual features
and textual tags whilst totally ignoring tag hierarchy information.
In the following experiments, 
we start with a two-layer tag hierarchy,
then evaluate the effect of tag layer number on the model performance.}

{\color{black} For image clustering, our aim is to reveal the category groups of 
	the dominant scene or event presented in these web images,  
	given the ground-truth available in group metadata
        \cite{johnson2015love,hu2016learning}.} 
To evaluate the performance of different tag completion methods, 
we divided the full tag labels into two parts:
observed part ($60\%$) with the remaining ($40\%$) as ground truth
\cite{feng2014image}.
The observed tags were randomly chosen.

\vspace{0.1cm}
\noindent \textbf{Visual features}:
For TRECVID MED 2011, we used HOG3D features \cite{klaser2008spatio} as visual representation of videos.
In particular, we first generated a codebook of $1000$ words using K-means \cite{jain2010data}.
With this codebook, 
we created a $1000$-D histogram feature vector for each video. 
Finally, the approximated Histogram Intersection Kernel via feature extension \cite{vedaldi2012efficient} 
was adopted to further enhance the expressive capability of visual features.
\zhu{
For NUS-WIDE, we exploited a VGG-16 convolutional neural network (CNN) \cite{simonyan2015very} pre-trained on 
the ImageNet Large-Scale Visual Recognition Challenge 2012 dataset \cite{russakovsky2015imagenet} to extract image features. 
This allows the image description benefiting from auxiliary rich object image annotation.
Specifically, we used the output (4096-D feature vector) 
from the first Fully-Connected CNN layer as image feature representation.}

\vspace{0.1cm}
\noindent \textbf{Implementation details}:
The default parameter settings are as follows. 
The forest size $\tau$ was fixed to $1000$
for all random forest models.
The depth of each tree was automatically determined 
by setting the sample number in the leaf node, $\phi$, 
which we set to $3$. 
We set $\nu_\mathrm{try} = \sqrt{d}$
with $d$ the data feature dimension (Equation~(\ref{eqn:split_parameter_optimisation}))
and $\kappa = 20$ (Equation \ref{eqn:tag_cpl_AM}).
For fair comparison, we used the exactly same number of clusters, 
visual features and tag data in all compared methods.
For any random forest model, we repeated $10$ folds and reported the average results.
In addition to the default settings above, we also evaluated the influence of two
important HML-RF parameters, e.g. $\tau$ and $\phi$ (Section \ref{sec:exp_model_analysis}).

\subsection{Evaluation on Discovering Global Data Cluster Structure}
\label{sec:evaluate_clustering}

\noindent \textbf{Input data modes}:
For comparison, we tested four modes of input data: 
(1) ViFeat: videos are represented by HOG3D visual features;
(2) BiTag: binary tag vectors are used instead of visual features;
(3) DetScore \cite{vahdat2014discovering}:  
tag classifiers (e.g. SVM) are trained for individual tags using the available tags with visual features and their detection scores are then used as model input\footnote{
	We only compared the reported results in \cite{vahdat2014discovering}	
	since we cannot reproduce the exact evaluation setting due to 
	the lack of experimental details.};
(4) ViFeat\&BiTag: both visual and tag data are utilised.
More specifically, the two modalities may be combined into one
single feature vector (called ViFeat\&BiTag-cmb),
or modelled separately in some balanced way (called ViFeat\&BiTag-bln),
depending on the design nature of specific methods.

\vspace{0.1cm}
\noindent \textbf{Baseline models}: 
We extensively compared our HML-RF model against 
the following related state-of-the-art methods:
(1) K-means \cite{jain2010data}:
The most popular clustering algorithm.
(2) Spectral Clustering (SpClust) \cite{ng2002spectral}: 
A popular and robust clustering mechanism based on the eigen-vector structures of affinity matrix.
\textcolor{black}{In ViFeat\&BiTag mode,
	the averaging over separate normalised affinity matrices of visual and
	tag data (SpClust-bln) was also evaluated, in addition to the combined single feature (SpClust-cmb). }
(3) Affinity Propagation (AffProp) \cite{frey2007clustering}:
An exemplar based clustering algorithm whose input is also affinity matrix.
This method is shown insensitive to exemplar initialisation as 
all data samples are simultaneously considered as potential cluster centres. 
(4) Clustering Random Forest (ClustRF) \cite{BreimanML01,shi2006unsupervised}:
A feature selection driven data similarity computing model. 
It was used to generate the data affinity matrix, followed by
SpClust for obtaining the final clusters.
(5) Constrained-Clustering Forest (CC-Forest) \cite{zhu2013video}:
A state-of-the-art multi-modality data based clustering forest characterised by 
joint learning of heterogeneous data. 
Its output is affinity matrix induced from all data modalities.
Similarly, the clusters are generated by SpClust.
(6) Affinity Aggregation for Spectral Clustering (AASC) \cite{huang2012affinity}:
A state-of-the-art multi-modal spectral clustering method that searches for an 
optimal weighted combination of multiple affinity matrices, each from a single data modality.
(7) CCA+SpClust \cite{hardoon2004canonical}:
The popular Canonical Correlation Analysis (CCA) model 
that maps two views (e.g. visual and tag features) to a common latent space
with the objective of maximising the correlation between the two. 
In this common space, we computed pairwise similarity between samples and  
applied the spectral clustering algorithm to obtain clusters. 
(8) 3VCCA+SpClust \cite{gong2014multi}:
A contemporary three-view CCA algorithm extended from the conventional CCA by
additionally considering the third view about high-level semantics.
Specifically, we utilised the first layer of abstract tags as the data of third view.
Similarly, we used spectral clustering on the similarity measures in the 
induced common space for data clustering.
(9) Maximum Margin Clustering (MMC) \cite{xu2004maximum}:
A widely used clustering model based on maximising the margin between clusters.
(10) Latent Maximum Margin Clustering (L-MMC) \cite{zhou2013latent}:
An extended MMC model that allows to accommodate latent variables, e.g. tag labels, during
maximum cluster margin learning. 
(11) Structural MMC (S-MMC) \cite{vahdat2014discovering}:
A variant of MMC model assuming structured tags are labelled on data samples.
(12) Flip MMC (F-MMC) \cite{vahdat2014discovering}:
The state-of-the-art tag based video clustering method capable of 
handling the missing tag problem, beyond S-MMC.
\zhu{
(13) Deep Canonical Correlation Analysis (DCCA) \cite{andrew2013deep}:
a deep neural network (DNN) based extension of CCA \cite{hardoon2004canonical}
where a separate DNN is used for extracting features of each data modality,
followed by canonical correlation maximisation between across-modal features.
(14) Deep Canonically Correlated Autoencoders (DCCAE) \cite{wang2015deep}:
a state-of-the-art deep multi-view learning method 
that combines the reconstruction errors of split autoencoder \cite{ngiam2011multimodal} and 
the correlation maximisation of DCCA \cite{andrew2013deep}
in model formulation.
}

\vspace{0.1cm}
\noindent \textbf{Evaluation metrics}:
We adopted five 
metrics to evaluate the clustering accuracy: 
(1) \textit{Purity} \cite{zhou2013latent}, 
which calculates the averaged accuracy of the dominating class in each cluster;
(2) \textit{Normalised Mutual Information} (NMI) \cite{vinh2009information},
which considers the mutual dependence between the predicted and ground-truth partitions;
(3) \textit{Rand Index} (RI) \cite{rand1971objective}, 
which measures the ratio of agreement between two partitions, 
i.e. true positives within clusters and true negatives between clusters;
(4) \textit{Adjusted Rand Index} (ARI) \cite{steinley2004properties}, 
an adjusted form of RI that additionally considers disagreement,
and equals $0$ when the RI equals its expected value; 
(5) \textit{Balanced F1 score} (F1) \cite{jardine1971use}, 
which uniformly measures both precision and recall. 
All metrics lie in the range of $[0, 1]$ except ARI in $[-1, 1]$.  
For each metric, higher values indicate better performance.
Whilst there may exist some inconsistency between different metrics
due to their property discrepancy \cite{vinh2010information},
using all them allows to various aspects of performance measure.

\begin{table} [h] \footnotesize 
  \centering
  \caption{
	  Comparing clustering methods
	  on TRECVID MED 2011 \cite{over2011trecvid}.
	  }
\label{tab:compare_clust_trecvid}
\vspace{0.2cm}
\renewcommand{\arraystretch}{1}
\setlength{\tabcolsep}{0.12cm}
    \begin{tabular}
    	{c||c|ccccc}
    \hline
    Input mode &{Method} & Purity & NMI & RI & F1 & ARI \\
    \hline \hline
    \multirow{5}[0]{*}{\rotatebox{0}{ViFeat}} 
    &{K-means\cite{jain2010data}} & 0.26  & 0.19  & 0.88  & 0.14  & 0.08 \\
    &{SpClust\cite{ng2002spectral}} & 0.25  & 0.20  & 0.88  & 0.15  & 0.07 \\    
    &{ClustRF\cite{BreimanML01}} & 0.23  & 0.17  & 0.87  & 0.14  & 0.08  \\
    &{AffProp\cite{frey2007clustering}} &  0.23   &  0.16     &      0.87 &   0.14    &  0.07 \\   
    &{MMC\cite{xu2004maximum}} & 0.25  & 0.19  & 0.88  & 0.14  & 0.09 \\
    \hline \hline
    \multirow{5}[0]{*}{\rotatebox{0}{BiTag}}
    &{K-means\cite{jain2010data}} & 0.51  & 0.52  & 0.86  & 0.30  & 0.23 \\
    &{SpClust\cite{ng2002spectral}} & 0.71  & 0.73  & 0.93  & 0.56  & 0.60 \\
    &{ClustRF\cite{BreimanML01}} & 0.77  & 0.81  & 0.94  & 0.64  & 0.60  \\
    &{AffProp\cite{frey2007clustering}} &  0.50    & 0.44  &  0.87   & 0.28      &  0.21\\  
    &{MMC\cite{xu2004maximum}} & 0.76  & 0.72  & 0.95  & 0.64  & 0.60 \\
    \hline\hline
    \multirow{4}[0]{*}{\rotatebox{0}{DetScore}}
    &{K-means\cite{jain2010data}} & 0.63  & 0.60  & 0.93  & 0.50  & - \\
    &{SpClust\cite{ng2002spectral}} & 0.82  & 0.76  & 0.96  & 0.69  & - \\
    &{MMC\cite{xu2004maximum}} & 0.83  & 0.78  & 0.96  & 0.73  & - \\
    &{L-MMC\cite{zhou2013latent}} & 0.86  & 0.82  & 0.97  & 0.79  & - \\
    \hline   \hline
    \multirow{1}[0]{*}{\rotatebox{0}{ViFeat\&}}
    &{K-means\cite{jain2010data}} & 0.51  & 0.49  & 0.90  & 0.34  & 0.24  \\
    BiTag-cmb&{SpClust-cmb\cite{ng2002spectral}} & 0.76  & 0.74  & 0.94  & 0.62  & 0.66  \\
    &{ClustRF\cite{BreimanML01}} & 0.23  & 0.17  & 0.87  & 0.15  & 0.08  \\
    &{AffProp\cite{frey2007clustering}} & 0.51  & 0.46  & 0.86  & 0.29  & 0.21  \\
    \hline
	\multirow{1}[0]{*}{\rotatebox{0}{ViFeat\&}}
	&{SpClust-bln\cite{ng2002spectral}} & 
    {0.75} & {0.72} & {0.95} & {0.62} & {0.59} \\
 
	BiTag-bln&   {CCA+SpClust\cite{hardoon2004canonical} }& {0.85} & {0.81} & {0.97} & {0.77} & {0.75} \\
    &  {3VCCA+SpClust\cite{gong2014multi}} & {0.86} & {0.86} & {0.97} & {0.78} & {0.77} \\    
    &{CC-Forest\cite{zhu2013video}} & 0.41  & 0.33  & 0.89  & 0.41 & 0.19  \\
    &{AASC\cite{huang2012affinity}} & 0.30  & 0.15  & 0.87  & 0.13  & 0.06  \\
   
    &{MMC\cite{xu2004maximum}} & 0.79  & 0.72  & 0.95  & 0.66  & 0.66  \\
    &{\zhu{DCCA\cite{andrew2013deep}}} & 0.84  & 0.80   & 0.96  & 0.74  & 0.72 \\
    &{\zhu{DCCAE\cite{wang2015deep}}} & 0.84  & 0.80   & 0.97  & 0.75  & 0.73 \\
    &{S-MMC\cite{vahdat2014discovering}} & 0.87  & 0.84  & 0.97  & 0.79  & - \\
    &{F-MMC\cite{vahdat2014discovering}} & 0.90  & 0.88  & \textbf{0.98} & 0.84  & - \\ 
    
    &HML-RF(Ours) & \textbf{0.94} & \textbf{0.90} & \textbf{0.98} & \textbf{0.88} & \textbf{0.87} \\
    \hline
    \end{tabular}%
\end{table}%

\subsubsection{Clustering Evaluation on TRECVID MED 2011}
We evaluated the effectiveness of distinct models for tag-based video clustering, 
using the \textit{full} tag data along with visual features. 
The results are reported in Table \ref{tab:compare_clust_trecvid}.
With visual features alone, all clustering methods produce poor results,
e.g. the best NMI is 0.20, achieved by SpClust.
Whereas binary tag representations provide much more information about the underlying 
video data structure than visual feature modality, 
e.g. all models can double their scores or even more in most metrics.
Interestingly, using the detection scores can lead to even better results than the original binary tags.
The plausible reason is that
missing tags can be partially recovered after using the detection scores.
When using both data modalities, 
we observed superior results than either single modality with many methods
like SpClust, AffProp, MMC.
This confirms the overall benefits from jointly learning visual and tag data 
because of their complementary effect.
Also, it is shown that separate and balanced use of visual and tag features (ViFeat\&BiTag-bln) is more likely to surpass methods using concatenated visual and tag vectors (ViFeat\&BiTag-cmb). 
A possible reason is that visual and tag features are heterogeneous 
to each other, a direct combination leads to an unnatural and inconsistent data representation
thus likely increases the modelling difficulty and deteriorates the model performance.

For the performance of individual methods,
the proposed HML-RF model evidently provides the best results by a significant margin over the second best Flip MMC in most metrics, 
except RI which is a less-sensitive measure due to its practical narrower range 
\cite{vinh2010information}. 
This is resulted from the joint exploitation of 
interactions between visual and tag data, 
tag hierarchical structure, and tag correlations with a unified HML-RF model
(Algorithm \ref{Alg:learn_HMLRF}),
different from MMC and its variants wherein
tags are exploited in a flat organisation and 
no tag dependences are considered.
K-means hardly benefits from visual and tag combination, 
due to its single distance function based grouping mechanism therefore
is very restricted in jointly exploiting multi-modal data.

Among all affinity based models, 
ClustRF 
is surprisingly dominated by visual data
when using visual features \& tag as input. 
This may be because that visual features with large variances may be mistakenly considered as optimum 
due to larger information gain induced on them.
CC-Forest suffers less by separately exploiting the two modalities,
but still inferior than HML-RF due to ignoring the intrinsic tag structure and 
the tag sparseness challenge.
AASC yields much poorer clustering results than HML-RF, 
suggesting 
that the construction of individual affinity matrices 
can lose significant information, such as the interactions
between the visual and tag data, as well as statistical tag correlations.

The methods of AffProp and SpClust-cmb also suffer from the heteroscedasticity problem 
in that the input affinity matrix is constructed 
from the heterogeneous concatenation of visual and tag data and thus
ineffective to exploit the knowledge embedded across modalities and tag statistical relationships.  
However, separating visual and tag features does not bring benefit to
SpClust (SpClust-bln). This may be due to tag sparseness and the lack of 
correlation modelling between visual and tag data.
\zhu{Whilst through correlating and optimising cross-modal latent common space, 
	correlation analysis models (e.g. CCA, DCCA, DCCAE and 3VCCA) overcome somewhat 
the heterogeneous data learning challenge
but remain suboptimal and inferior due to over-sparse tags 
and the ignorance of tag hierarchy and
inter-tag correlations.}

\begin{table} [h] \footnotesize 
	\centering
	\caption{
		Comparing clustering methods on NUS-WIDE \cite{nus-wide-civr09}.
	}
	\label{tab:cluster_nus}
	\vspace{0.2cm}
	\renewcommand{\arraystretch}{1}
	\setlength{\tabcolsep}{0.12cm}
	\begin{tabular}
		{c||c|ccccc}
		\hline
		Input mode &{Method} & Purity & NMI & RI & F1 & ARI \\
		\hline \hline
		\multirow{5}[0]{*}{\rotatebox{0}{ViFeat}} 
		&{K-means\cite{jain2010data}}  & 0.28  & 0.26  & 0.94  & 0.13  & 0.11 \\
		&{SpClust\cite{ng2002spectral}} & 0.27  & 0.24  & 0.94  & 0.14  & 0.11 \\
		&{ClustRF\cite{BreimanML01}} & 0.27  & 0.24  & 0.94  & 0.14  & 0.11 \\
		&{AffProp\cite{frey2007clustering}} & 0.25  & 0.22  & 0.91  & 0.13  & 0.09 \\
		&{MMC\cite{xu2004maximum}} & 0.24  & 0.20   & 0.94  & 0.12  & 0.09 \\
		\hline \hline
		\multirow{5}[0]{*}{\rotatebox{0}{BiTag}}
		&{K-means\cite{jain2010data}}  & 0.46  & 0.64  & 0.77  & 0.20   & 0.15 \\
		&{SpClust\cite{ng2002spectral}} & 0.51  & 0.59  & 0.72  & 0.12  & 0.06 \\
		&{ClustRF\cite{BreimanML01}}  & 0.57  & 0.60   & 0.90   & 0.15  & 0.33 \\
		&{AffProp\cite{frey2007clustering}}  & 0.50   & 0.59  & 0.76  & 0.15  & 0.16 \\
		&{MMC\cite{xu2004maximum}}  & 0.54  & 0.61  & 0.94  & 0.24  & 0.40 \\
		\hline\hline
		\multirow{4}[0]{*}{\rotatebox{0}{DetScore}}
		&{K-means\cite{jain2010data}} & 0.51  & 0.65  & 0.79  & 0.22  & 0.17 \\
		&{SpClust\cite{ng2002spectral}} & 0.55  & 0.61  & 0.75 & 0.16  & 0.09 \\
		&{ClustRF\cite{BreimanML01}}  & 0.60  & 0.62   & 0.92   & 0.17  & 0.35 \\
		&{AffProp\cite{frey2007clustering}}  & 0.54   & 0.60  & 0.78  & 0.17  & 0.18 \\
		&{MMC\cite{xu2004maximum}} & 0.59  & 0.64  & 0.95  & 0.25  & 0.41 \\		
		\hline   \hline
		\multirow{1}[0]{*}{\rotatebox{0}{ViFeat\&}}
		&{K-means\cite{jain2010data}}  & 0.29  & 0.26  & 0.94  & 0.14  & 0.11 \\
		BiTag-cmb&{SpClust-cmb\cite{ng2002spectral}} & 0.28  & 0.24  & 0.94  & 0.14  & 0.11 \\
		&{ClustRF\cite{BreimanML01}} & 0.28  & 0.24  & 0.93  & 0.13  & 0.09 \\
		&{AffProp\cite{frey2007clustering}}  & 0.26  & 0.22  & 0.91  & 0.13  & 0.10 \\
		\hline
		\multirow{1}[0]{*}{\rotatebox{0}{ViFeat\&}}
		&{SpClust-bln\cite{ng2002spectral}} & 0.58  & 0.56  & 0.87  & 0.19  & 0.14 \\
		
		BiTag-bln&   {CCA+SpClust\cite{hardoon2004canonical} }  & 0.48  & 0.41  & 0.95  & 0.24  & 0.28 \\
		&  {3VCCA+SpClust\cite{gong2014multi}}   & 0.52  & 0.45  & \textbf{0.96}  & 0.25  & 0.32 \\
		&{CC-Forest\cite{zhu2013video}}  & 0.26  & 0.23  & 0.91  & 0.12  & 0.07 \\
		&{AASC\cite{huang2012affinity}}  & 0.28  & 0.24  & 0.94  & 0.13  & 0.10 \\
		
		&{MMC\cite{xu2004maximum}}& 0.24  & 0.20   & 0.94  & 0.12  & 0.09 \\
		&{DCCA\cite{andrew2013deep}} & 0.61  & 0.62  & 0.89  & 0.30   & 0.27 \\
		&{DCCAE\cite{wang2015deep}} & 0.62  & 0.63  & 0.89  & 0.30   & 0.27 \\
		&HML-RF(Ours) & \textbf{0.67} & \textbf{0.67} & \textbf{0.96} & \textbf{0.32} & \textbf{0.45} \\
		\hline
	\end{tabular}%
\end{table}%

\begin{figure*} 
	\centering
	\includegraphics[width=0.99\linewidth]{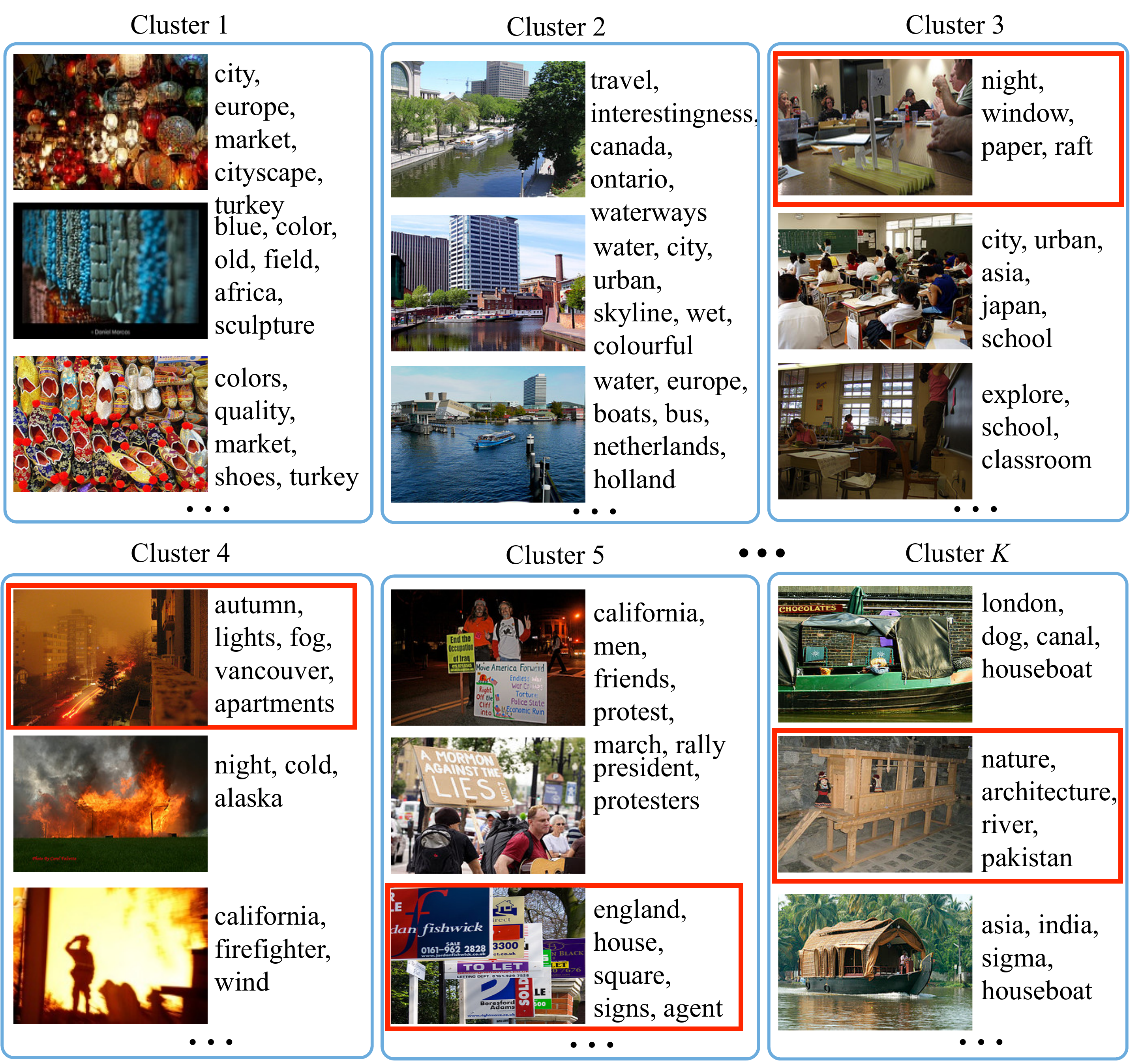}
	\vskip -.3cm
	\caption{
		Example clusters discovered by the HML-RF model on NUS-WIDE \cite{nus-wide-civr09}.
		Tags are shown at the right of corresponding images. 
		The inconsistent samples are indicated with red bounding box.
	}
	\label{fig:example_clust_NUS}
\end{figure*}

\subsubsection{Clustering Evaluation on NUS-WIDE}
\label{sec:clustering_NUS}
We further evaluated the proposed HML-RF model and its competitors on tagged image dataset NUS-WIDE \cite{nus-wide-civr09}. 
\zhu{In this experiment, we utilised a two-layer tag hierarchy in HML-RF.} 
The clustering results are reported in Table \ref{tab:cluster_nus}.
It is evident that our HML-RF model surpasses all baseline methods, consistent with the findings in clustering TRECVID videos.
Specifically, methods based on SpClust obtain generally
more accurate clusters. 
Interestingly, simple combination of affinity matrices
(SpClust-bln) is shown superior than latent common subspace learning (CCA and 3VCCA).
This is opposite from the observations on the TRECVID videos above.
A possible explanation may be due to 
the additional difficulty for joint subspace learning caused by
the greater tag sparseness on NUS-WIDE images,
e.g. missing tags making the learned projection inaccurate and suboptimal. 
\zhu{
Deep leaning based DCCA and DCCAE methods also suffer from the same problem although
their stronger modelling capability can improve considerably the quality of learned subspaces.
By incorporating tag hierarchy knowledge and employing automatically
mined tag correlations, our HML-RF model mitigates more effectively such
tag sparsity and incomplete cross-modal data alignment challenges.}
This again suggests the capability and effectiveness of our method in 
exploiting sparse tags for discovering global visual data concept structure.
Example of image clusters discovered by our HML-RF are shown in Figure \ref{fig:example_clust_NUS}. 
%

\subsubsection{Further Analysis}
\label{sec:exp_model_analysis}
We further conducted a series of in-depth evaluations and analysis: 
(1) model robustness against tag sparseness;  
(2) HML-RF model component effect; 
(3) HML-RF model parameter sensitivity;
and 
\zhu{(4) tag hierarchy structure effect}.

\begin{figure} 
	\centering
	\includegraphics[width=0.9\linewidth]{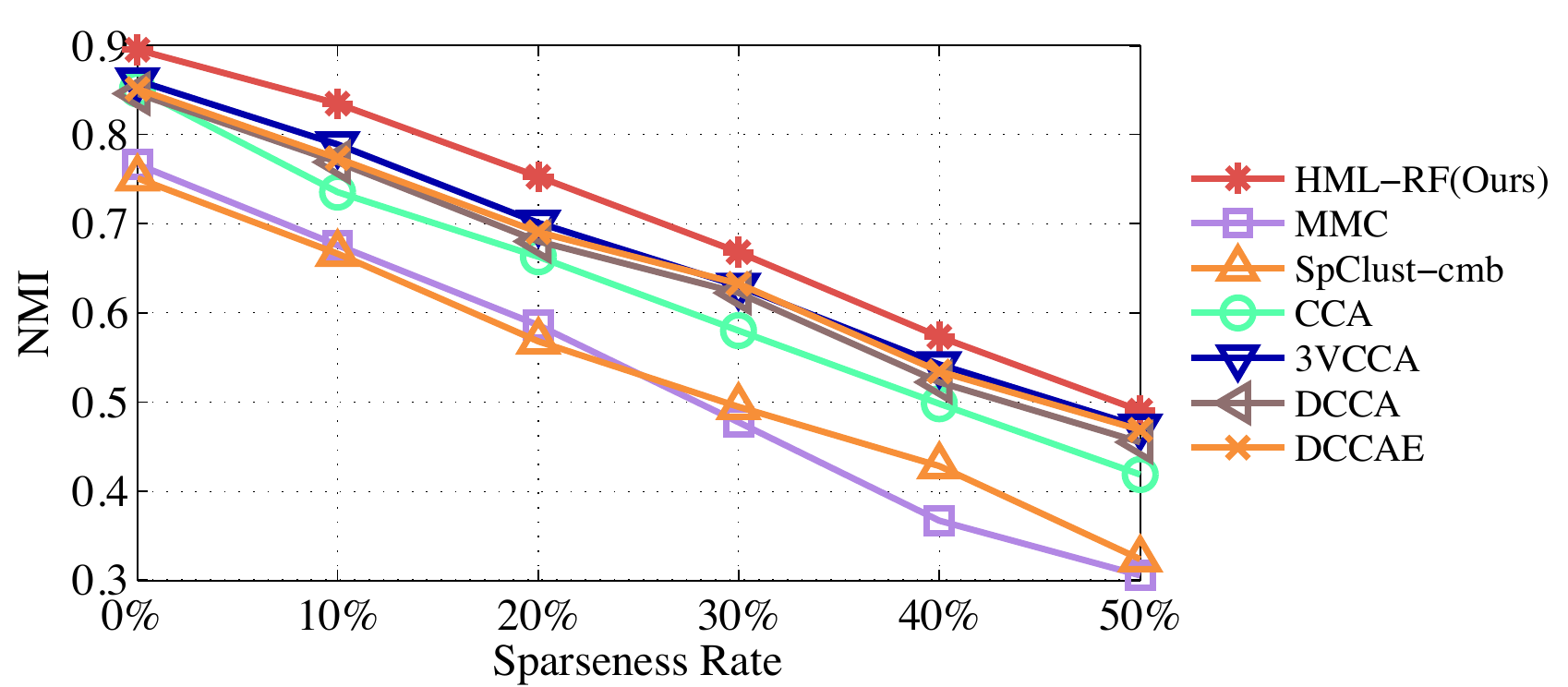}
	\vskip -.5cm
	\caption{
		Clustering performance in NMI of compared methods at different tag sparseness rates on TRECVID MED 2011 \cite{over2011trecvid}.
	} 
	\label{fig:sparse}
\end{figure}

\begin{figure} 
	\centering
	\includegraphics[width=.88\linewidth]{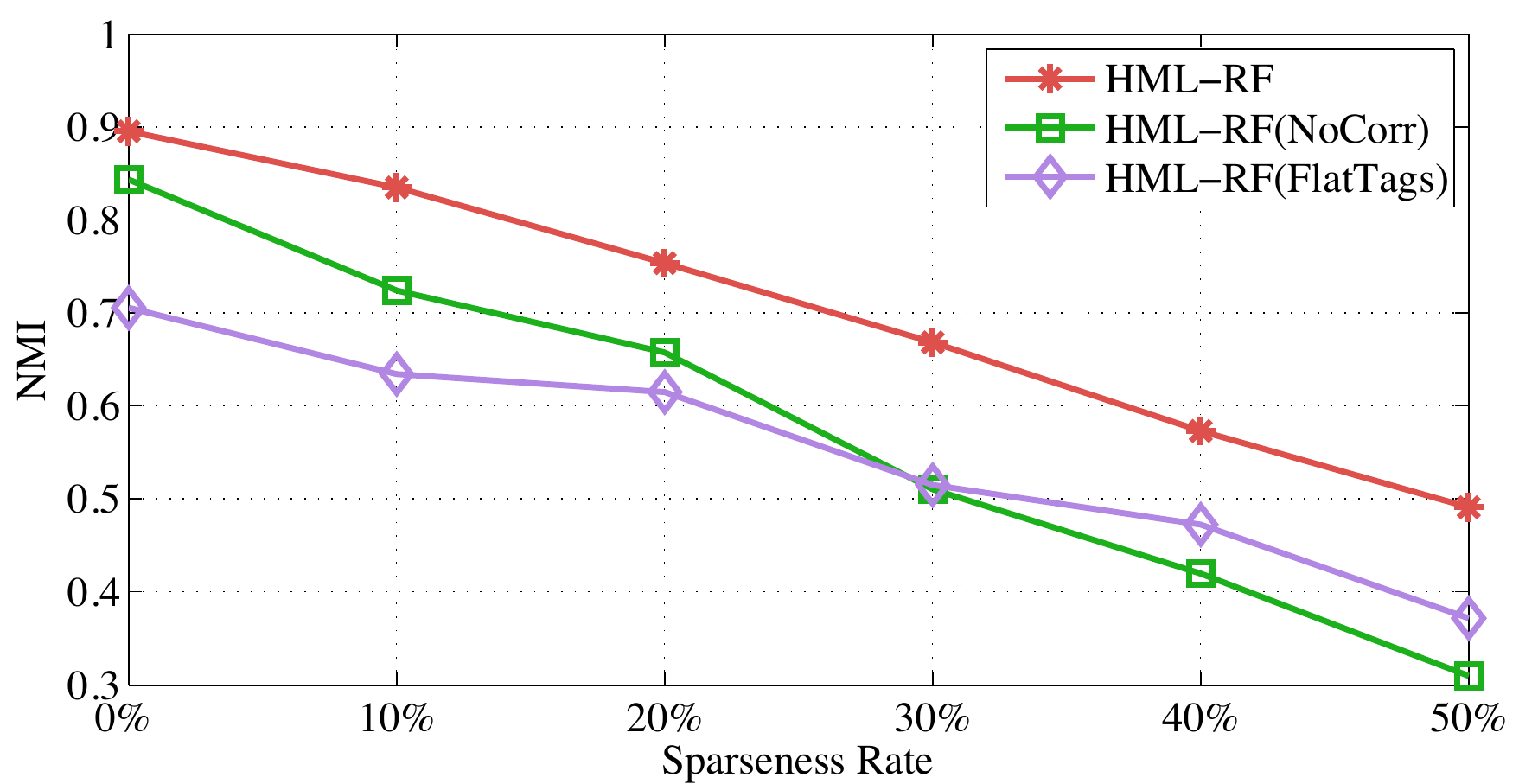}
	\vskip -0.3cm
	\caption{
		Evaluating the effectiveness of specific HML-RF components on TRECVID MED 2011 \cite{over2011trecvid}.
	} 
	\label{fig:HMLRFs}
	
\end{figure}

\vspace{0.1cm}
\noindent \textbf{Model robustness against tag sparseness}:
We conducted a scalability evaluation against tag sparseness and incompleteness.
This is significant since we may have access to merely a small size of tags in many practical settings.
To simulate these scenarios, 
\zhu{we randomly removed varying ratios ($10\% \sim 50\%$) of tag data} on the TRECVID MED 2011 dataset.
%
We utilised both visual and tag data as model input 
since most methods can benefit from using both\footnote{
	Structural MMC and Flip MMC models \cite{vahdat2014discovering} were not included in this evaluation 
	due to the difficulties in reproducing their models from a lack of
	sufficient implementation details.}.
The most common metric NMI \cite{jain2010data} was used
in this experiment.

\begin{figure*} 
	\centering
	\includegraphics[width=0.97\linewidth]{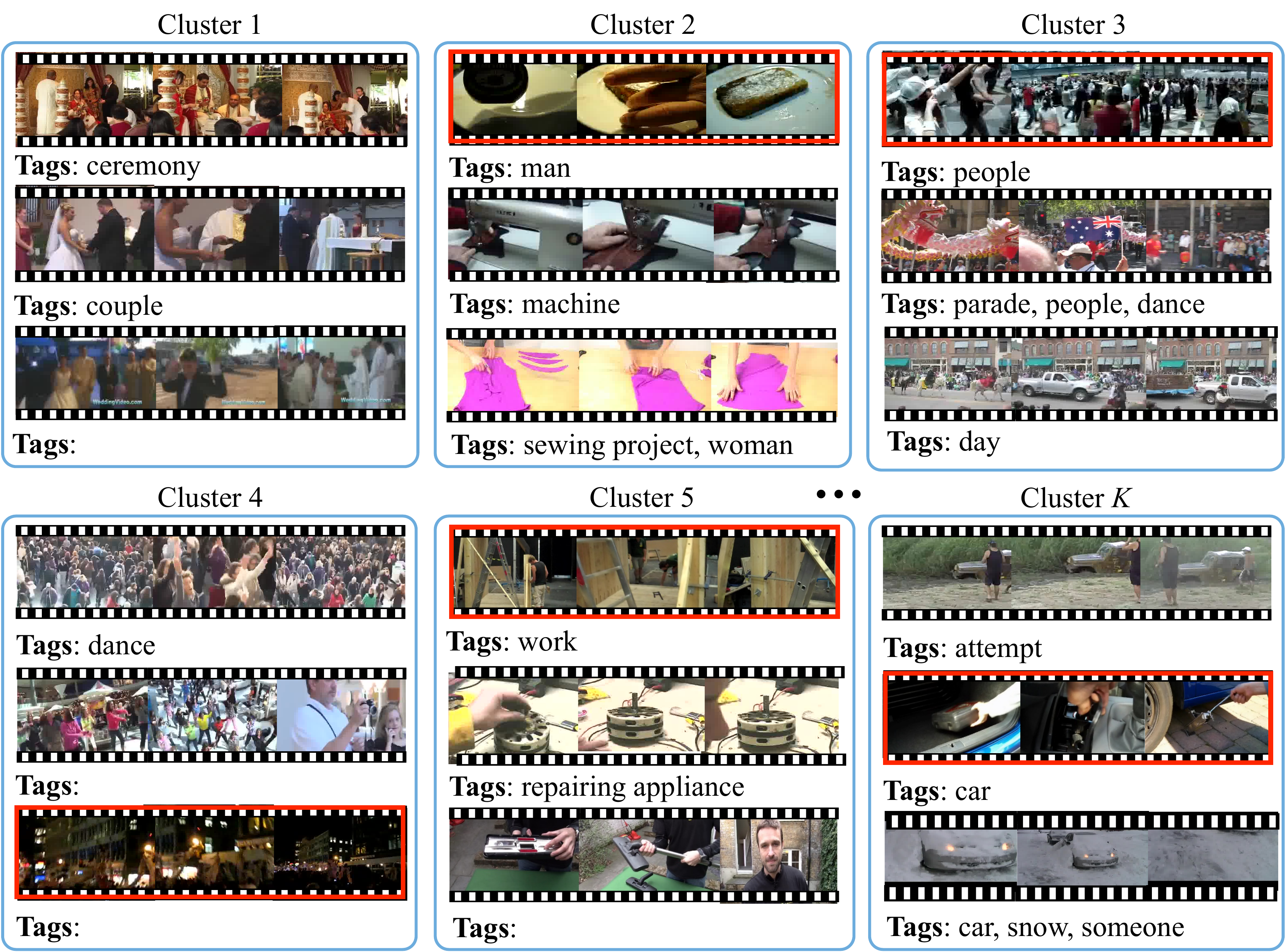}
	\vskip -0.2cm
	\caption{
		\zhu{
			Example clusters formed by the HML-RF model given $50\%$ tag sparseness rate 
			on TRECVID MED 2011 \cite{over2011trecvid}. 
			Tags are shown underneath the corresponding video. 
			Not that some videos have no tag data.
			Inconsistent samples are indicated with red bounding box.}
	}
	\label{fig:example_clust_TRECVID}
\end{figure*}

\begin{table} 
	\footnotesize
	\centering
	\caption{
		Comparing relative drop in NMI of top-$7$ clustering models,
		given different tag sparseness rates on TRECVID MED 2011 \cite{over2011trecvid}.
		Smaller is better.
	}
	\renewcommand{\arraystretch}{1}
	\vspace{0.2cm}
	\label{tab:rel_drop_NMI}%
	\footnotesize
	\begin{tabular}{r|cccccc}
		\hline
		Sparseness rate (\%) & 10   & 20  & 30  & 40  & 50 \\
		\hline
		SpClust-cmb\cite{ng2002spectral}  
		& 0.11  & 0.24  & 0.34  & 0.43  & 0.57  \\
		MMC\cite{xu2004maximum}    & 0.12  & 0.24  & 0.38  & 0.52  & 0.60  \\
		CCA+SpClust\cite{hardoon2004canonical}
		& 0.14  & 0.22  & 0.32  & 0.41  & 0.51  \\
		3VCCA+SpClust\cite{gong2014multi} 
		& 0.08  & 0.19  & 0.27  & 0.37  & \textbf{0.45} \\
		\zhu{DCCA}\cite{hardoon2004canonical}
		& 0.09  & 0.20  & 0.26  & 0.38  & 0.46  \\
		\zhu{DCCAE}\cite{gong2014multi} 
		& 0.09  & 0.19  & 0.26  & 0.37  & \textbf{0.45} \\
		HML-RF(Ours) &  \textbf{0.07} & \textbf{0.16} & \textbf{0.25} & \textbf{0.36} & \textbf{0.45} \\
		\hline
	\end{tabular}%
\end{table}%

\begin{figure*} 
	\centering
	\includegraphics[width=0.45\linewidth]{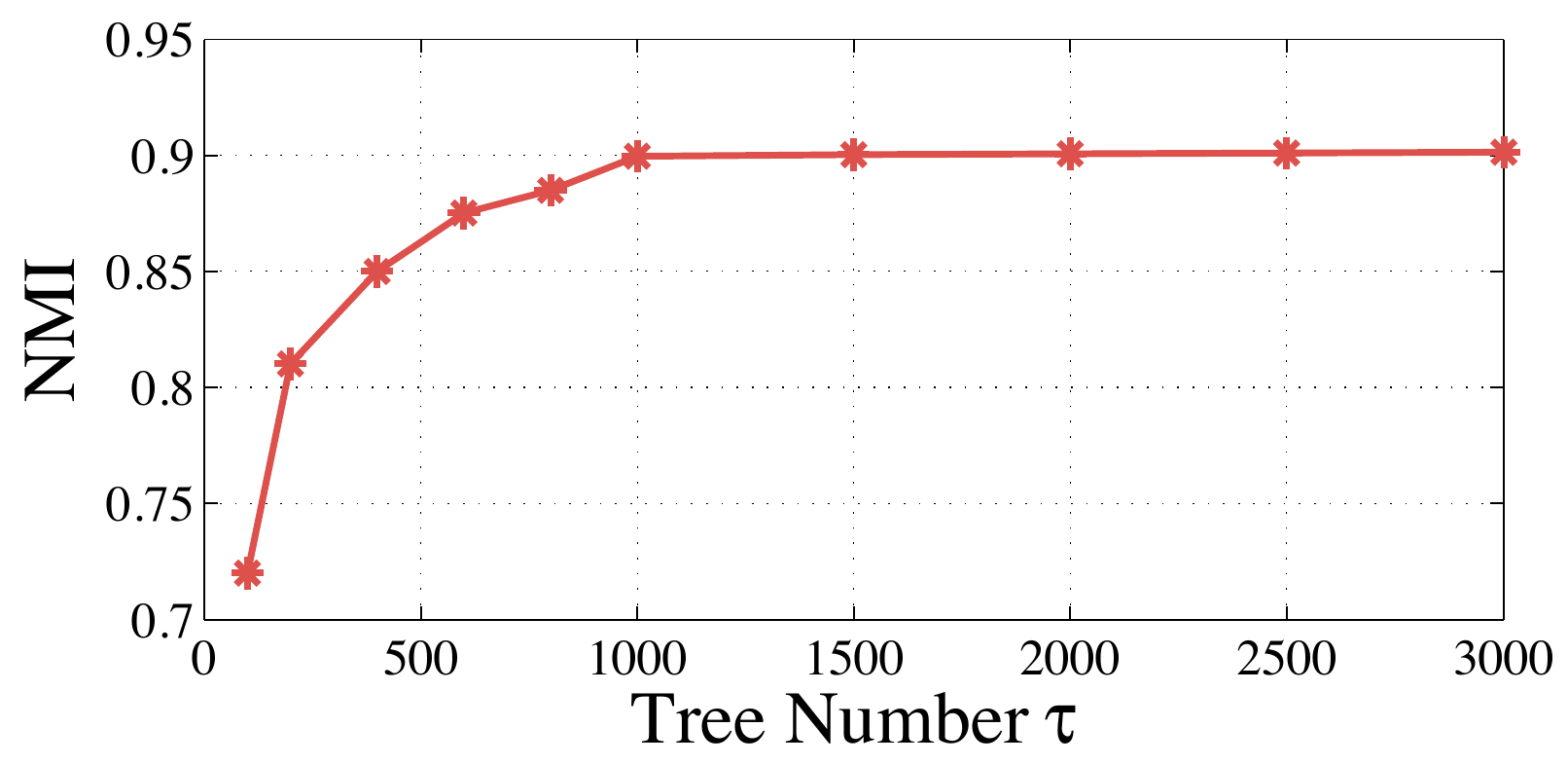}
	\includegraphics[width=0.45\linewidth]{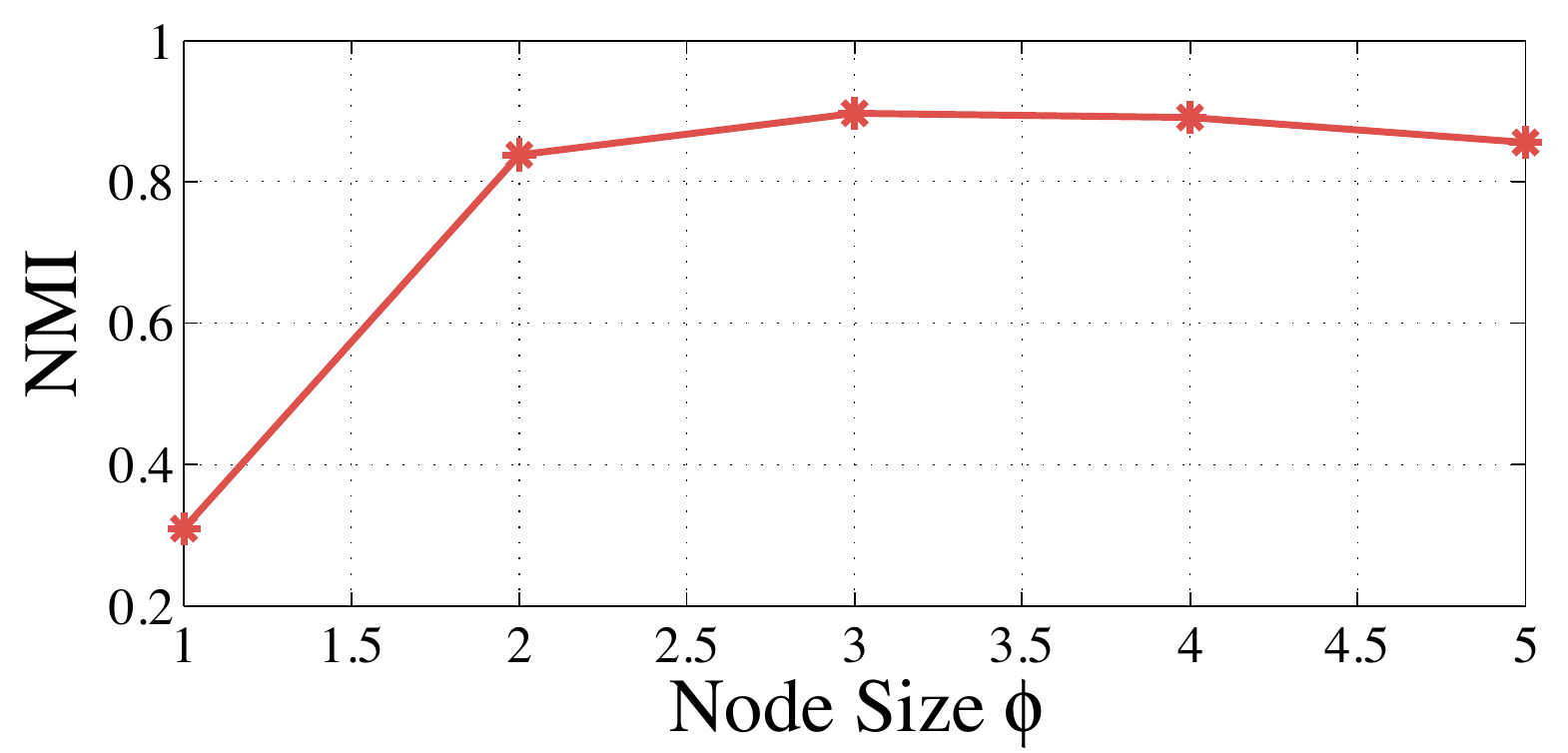}
	\vskip -0.3cm
	\caption{
		Clustering performance in NMI of HML-RF over different forest sizes ($\tau$) and node size ($\phi$) on TRECVID MED 2011 \cite{over2011trecvid}.
	}
	\label{fig:tree_size}
\end{figure*}

The results by top-7 clustering methods are compared in Figure \ref{fig:sparse}.
Given less amount of tag data, 
as expected we observe a clear performance drop trend across all these models. 
{\color{black} However, the relative drops in the performance of HML-RF model
  due to tag incompleteness 
	are the smallest among all compared methods at $10\%\sim40\%$
        sparseness rate (less is more sparse). This performance
        degradation is comparable among three best models (HML-RF,
        3VCCA and DCCAE) at $50\%$ sparseness rate, as shown in Table
        \ref{tab:rel_drop_NMI}.}
This demonstrates the robustness and benefits of the proposed HML-RF model 
with respect to tag sparseness and incompleteness, and making it more practically
useful when fewer tags are available.
This also demonstrates that a joint exploitation of visual features, tags hierarchy as well as tag correlations can bring about significant benefits to visual semantic structure interpretation and global video clustering with sparse/incomplete tags.
For qualitative visualisation, an example of clusters formed by our HML-RF under the most sparse case is given
in Figure \ref{fig:example_clust_TRECVID}.

\vspace{0.1cm}
\noindent \textbf{HML-RF model component effect}: 
We explicitly examined two components of the proposed HML-RF for
casting light on model formulation:
(1) the effect of exploiting tag abstractness hierarchy structure; and 
(2) the influence of tag statistical correlations.
To that end, we build two stripped-down variants of HML-RF:
{\bf(I)} HML-RF(FlatTags): A HML-RF without exploiting tag hierarchy and tag correlations (Equation (\ref{eqn:our_infogain_multitag}));
{\bf (II)} HML-RF(NoCorr): A HML-RF without tag correlation (Equation (\ref{eqn:our_infogain_hierarchy})).
Contrasting the performance between HML-RF(FlatTags) and HML-RF(NoCorr) allows for measuring the former,
whilst that between HML-RF(NoCorr) and HML-RF for the later.
We repeated the same experiments as above with the two variants.
\zhu{It is evident from Figure \ref{fig:HMLRFs} 
that both components make significant differences 
but their relative contribution varies under different tag sparseness cases.
Particularly, given the full tags, 
tag abstractness hierarchy plays a significant role, 
e.g. boosting NMI from $0.71$ to $0.84$;
but when more sparse tag data is utilised, 
the performance gain decreases and even drops at $>30\%$ sparseness rates. 
However, combining with tag correlations can effectively increase
the clustering accuracy.
This indicates that the tag hierarchy component works under certain tag densities
and coordinates well with tag correlations particularly in sparse tag cases.
On the other hand, an opposite phenomenon takes place with tag correlations,
i.e. it brings large benefit 
(from $0.31$ to $0.49$) 
in the most sparse case.}
These observations suggest that the two components are complementary and both are
important constitutes of the unified HML-RF model.

\begin{table} 
	\footnotesize
	\renewcommand{\arraystretch}{1}
	\centering
	\caption{
		Evaluating the effect of tag hierarchy layer number in clustering performance 
		by our HML-RF model on NUS-WIDE \cite{nus-wide-civr09}. }
	\begin{tabular}{c|ccccc}
		\hline
		Tag layer number & Purity & NMI & RI & F1 & ARI \\
		\hline
		2  &  0.67  & 0.67  & \textbf{0.96} & 0.32  & 0.45 \\
		3  &  0.68  & 0.68  & \textbf{0.96} & 0.34  & 0.47 \\
		4  &  {0.70} & {0.69} & \textbf{0.96} & \textbf{0.35} & \textbf{0.48} \\
		5  &  
		\textbf{0.71} & \textbf{0.70} & \textbf{0.96} & \textbf{0.35} & \textbf{0.48} \\
		6  &  
		\textbf{0.71} & \textbf{0.70} & \textbf{0.96} & \textbf{0.35} & \textbf{0.48} \\
		7  &  
		\textbf{0.71} & \textbf{0.70} & \textbf{0.96} & \textbf{0.35} & \textbf{0.48} \\ 
		\hline
	\end{tabular}%
	\label{tab:multi-layer}%
\end{table}%

\begin{table*} 
	\footnotesize
	\color{black}
	\renewcommand{\arraystretch}{1}
	\centering
	\caption{
		Evaluating the effect of tag abstractness in the
		HML-RF model on data clustering. } 
	\vspace{0.2cm}
	\begin{tabular}{c||c|ccccc}
		\hline
		Dataset & Tag Structure & Purity & NMI & RI & F1 & ARI \\
		\hline  \hline
		\multirow{2}{*}{TRECVID MED 2011 \cite{over2011trecvid}} 
		&  1-Layer Most Specific Tags  &  0.39  & 0.33  & 0.88 & 0.24  & 0.18 \\
		\cline{2-7}
		& 2-Layers Hierarchy Tags & 
		\textbf{0.94} & \textbf{0.90} & \textbf{0.98} & \textbf{0.88} & \textbf{0.87} \\
		
		\hline \hline
		
		\multirow{2}{*}{NUS-WIDE \cite{nus-wide-civr09}} 
		& 1-Layer Most Specific Tags  &  0.25   & 0.24   &  0.92 &  0.11 & 0.07 \\
		\cline{2-7}
		& 2-Layers Hierarchy Tags & 
		\textbf{0.67} & \textbf{0.67} & \textbf{0.96} & \textbf{0.32} & \textbf{0.45} \\  
		\hline
	\end{tabular}%
	\label{tab:layer-effect}%
\end{table*}%

\begin{table*} 
	\footnotesize
	\centering  
	\caption{
		Comparing Precision \& Recall between tag completion methods on TRECVID MED 2011 \cite{over2011trecvid}.
	}
	\label{tab:AP_AR_TRECVID}%
	\renewcommand{\arraystretch}{1}
	\setlength{\tabcolsep}{0.2cm}
	\vspace{0.2cm}
	\begin{tabular}{c||ccccc||ccccc}
		\hline
		Metric & \multicolumn{5}{c||}{AP@$N$} & \multicolumn{5}{c}{AR@$N$} \\
		\hline
		Recovered tag \# $N$ 
		& 1     & 2     & 3     & 4     & 5 
		& 1     & 2     & 3     & 4     & 5 \\
		\hline \hline
		LSR \cite{lin2013image}  
		& 0.31  & 0.25  & 0.22  & 0.20  & 0.17 
		& 0.15  & 0.24  & 0.32  & 0.38  & 0.40 \\
		TCMR \cite{feng2014image}  
		& 0.35  & 0.27  & 0.24  & 0.22  & 0.20 
		& 0.17  & 0.26  & 0.35  & 0.43   & 0.48 \\
		\hline
		AASC(GC) 
		& 0.23  & 0.17  & 0.14  & 0.13  & 0.11 
		& 0.12  & 0.18  & 0.21  & 0.25  & 0.27 \\
		SpClust-bln(GC) 
		& 0.31  & 0.27  & 0.25  & 0.23  & 0.21
		& 0.15  & 0.25  & 0.36  & 0.44  & 0.48 \\
		CC-Forest(GC) 
		& 0.28  & 0.24  & 0.18  & 0.15  & 0.14
		& 0.15  & 0.23  & 0.26  & 0.27  & 0.31 \\
		CCA+SpClust(GC)  
		& 0.34  & 0.29  & 0.26  & 0.26  & 0.23
		& 0.16  & 0.26  & 0.34  & 0.47  & 0.52 \\
		3VCCA+SpClust(GC) 
		& 0.35  & 0.29  & 0.26  & 0.26  & 0.23
		& 0.17  & 0.27  & 0.34  & 0.47  & 0.52 \\
		MMC(GC) 
		& 0.32  & 0.25  & 0.23  & 0.24  & 0.21
		& 0.15  & 0.24  & 0.36  & 0.45  & 0.49 \\
		
		\zhu{DCCA(GC)}
		& 0.35  & 0.29  & 0.26  & 0.26  & 0.23
		& 0.17  & 0.27  & 0.34  & 0.47  & 0.52 \\
		\zhu{DCCAE(GC)}
		& 0.36  & 0.29  & \textbf{0.27}  & 0.26  & 0.24
		& 0.17  & 0.27  & 0.35  & 0.47  & 0.53 \\
		\hline
		HML-RF(GC) 
		& 0.36  & \textbf{0.31} & \textbf{0.27} & \textbf{0.27} & \textbf{0.25}
		& 0.17 & \textbf{0.29} & \textbf{0.37} & \textbf{0.49} & \textbf{0.56} \\
		
		HML-RF(LN)   
		& 0.37 & 0.29   & 0.25  & 0.23  & 0.20
		& \textbf{0.19} & 0.28  & 0.34  & 0.44  & 0.49 \\
		
		HML-RF(AM) 
		& {\bf 0.38} & 0.30   & 0.26  & 0.24  & 0.22
		& 0.18 & 0.27 & 0.36 & 0.44 & 0.50\\
		
		\hline
	\end{tabular}%
\end{table*}%

\vspace{0.1cm}
\noindent \textbf{HML-RF model parameter sensitivity}:
	We evaluated two key parameters in HML-RF: Tree number $\tau$
	and leaf node size $\phi$.
	The results are given in Figure \ref{fig:tree_size}.
	It is evident that when more trees are trained and utilised, the clustering accuracy increases monotonically and starts to converge from $\tau=1000$. 
	This is consistent with the findings in \cite{Criminisi2012,shotton2008semantic}.
	When $\phi = 1$, weaker clustering results are obtained.
	This makes sense because HML-trees are overly grown, 
	e.g. they enforce very similar data samples to be separated and thus make the pairwise affinity estimation inaccurate (Section \ref{sec:clustering}).
	Setting small values to $\phi$ significantly improves the
        clustering accuracy, and is shown to be insensitive w.r.t. specific numbers.

\vspace{0.1cm}
\noindent \zhu{\textbf{Tag hierarchy structure effect}:
	Apart form two-layer tag hierarchy, we further evaluated the effect of tag layer number on
	the clustering performance of our HML-RF model on the NUS-WIDE \cite{nus-wide-civr09} dataset. 
	Specifically, we evaluated different tag hierarchies ranging from $3$ to $7$ layers,
	and the results are shown in Table \ref{tab:multi-layer}.
	We made these observations:
	(1) The layer number of tag hierarchy can 
	affect the results of data structure discovery by our HML-RF model;
	(2) The NUS-WIDE tags may lie in multiple abstractness layers, 
	which leads to better discovered cluster structure than that by two layers; 
	(3) The performance starts to get saturated from five layers and appending further more layers has little effect on data structure discovery,
	probably due to that over specific tags have little influence on 
	data structure.
	These findings imply the effectiveness and robustness of HML-RF in
	accommodating tag hierarchies of various structures and qualities. 
}

\vspace{0.1cm}
\noindent \zhu{\color{black} \textbf{Tag abstractness effect}: 
We further evaluated the benefit of tag abstractness by comparing
(i) the 2-layers tag hierarchy structure with (ii) a 1-layer
structure of the most specific tags in the proposed HML-RF model.
Table \ref{tab:layer-effect} shows a significant performance advantage from
exploiting a hierarchical tag abstractness structure for 
data clustering on both the TRECVID MED 2011 and the NUS-WIDE datasets.
This demonstrates more clearly the effectiveness of HML-RF in mining
and exploiting semantic information from multiple levels of tag abstractness
for global data structure analysis.
}

\begin{table*} 
	\footnotesize
	\centering
	\caption{
		Comparing Coverage@$N$ between different tag completion methods.
	}
	\vspace{0.2cm}
	\renewcommand{\arraystretch}{1}
	\setlength{\tabcolsep}{0.15cm}
	\begin{tabular}{r||ccccc||ccccc}
		\hline
		Dataset & \multicolumn{5}{c||}{TRECVID MED 2011 \cite{over2011trecvid}} & \multicolumn{5}{c}{NUS-WIDE \cite{nus-wide-civr09}} \\
		\hline
		Recovered tag \# $N$ 
		& 1     & 2     & 3     & 4     & 5
		& 1     & 2     & 3     & 4     & 5 \\
		\hline \hline
		LSR \cite{lin2013image}   
		& 0.31  & 0.43  & 0.52  & 0.59  & 0.61 
		& 0.30  & 0.35  & 0.38  & 0.40  & 0.42 \\
		TCMR \cite{feng2014image} 
		& 0.35  & 0.46   & 0.57  & 0.66  & 0.71 
		& 0.25  & 0.33  & 0.39  & 0.43  & 0.46 \\
		\hline
		AASC(GC)
		& 0.23  & 0.33  & 0.38  & 0.43  & 0.46
		& 0.09  & 0.14  & 0.17  & 0.22  & 0.22 \\
		SpClust-bln(GC) & 0.31  & 0.44  & 0.55  & 0.63  & 0.65 
		& 0.15  & 0.21  & 0.25  & 0.29  & 0.33 \\
		CC-Forest(GC) & 0.28  & 0.43  & 0.47  & 0.48  & 0.52 
		& 0.08  & 0.13  & 0.17  & 0.21  & 0.21 \\
		CCA+SpClust(GC)   & 0.34  &0.45   &0.56  &0.65    &  0.70
		& 0.15  & 0.22  & 0.27  & 0.32  & 0.36 \\
		3VCCA+SpClust(GC)   & 0.35  &0.46   &0.56  &0.65    &  0.70 
		& 0.16  & 0.23  & 0.28  & 0.32  & 0.36 \\
		MMC(GC) & 0.32   & 0.42   &0.55    &0.66    &  0.70 
		& 0.10  & 0.15  & 0.18  & 0.24  & 0.23 \\
		\zhu{DCCA(GC)}
		& 0.35  &0.46   &0.56  &0.66    &  0.70
		& 0.18  & 0.21  & 0.27  & 0.29  & 0.33 \\
		\zhu{DCCAE(GC)}
		& 0.36  &0.47   &0.57  &0.66    &  0.71
		& 0.18  & 0.23  & 0.27  & 0.29  & 0.33 \\
		\hline
		HML-RF(GC) & 0.36  & \textbf{0.49} & \textbf{0.59} & \textbf{0.68} & \textbf{0.75} 
		& 0.20  & 0.26  & 0.30  & 0.32  & 0.35 \\
		HML-RF(LN) & 0.37 & \textbf{0.49} & 0.56  & 0.65  & 0.68 
		& 0.29  & 0.35  & 0.39  & 0.41  & 0.42 \\		
		HML-RF(AM)& \textbf{0.38}   &0.47   & 0.58  &  0.65 &   0.70
		& \textbf{0.34} & \textbf{0.41} & \textbf{0.45} & \textbf{0.48} & \textbf{0.50} \\
		\hline
	\end{tabular}%
	\label{tab:coverage} %
\end{table*}%

\begin{table*} 
	\footnotesize
	\centering  
	\caption{
		Comparing Precision \& Recall between different tag completion methods on NUS-WIDE \cite{nus-wide-civr09}.
	}
	\label{tab:AP_AR_NUS}%
	\renewcommand{\arraystretch}{1}
	\setlength{\tabcolsep}{0.15cm}
	\vspace{0.2cm}
	\begin{tabular}{r||ccccc||ccccc}
		\hline
		Metric & \multicolumn{5}{c||}{AP@$N$} & \multicolumn{5}{c}{AR@$N$} \\
		\hline
		Recovered tag \# $N$ 
		& 1     & 2     & 3     & 4     & 5 
		& 1     & 2     & 3     & 4     & 5 \\
		\hline \hline
		LSR \cite{lin2013image}  
		& 0.30  & 0.22  & 0.18  & 0.15  & 0.13
		& 0.15  & 0.21  & 0.24  & 0.27  & 0.28 \\
		TCMR \cite{feng2014image}  
		& 0.25  & 0.19  & 0.16  & 0.15  & 0.13   
		& 0.13  & 0.19  & 0.23  & 0.26  & 0.29 \\
		\hline
		AASC(GC) 
		& 0.09  & 0.09  & 0.09  & 0.07  & 0.07   
		& 0.05  & 0.07  & 0.10  & 0.12  & 0.15 \\
		SpClust-bln(GC) 
		& 0.15  & 0.12  & 0.10  & 0.08  & 0.09
		& 0.09  & 0.13  & 0.15  & 0.20  & 0.20 \\
		CC-Forest(GC) 
		& 0.08  & 0.09  & 0.09  & 0.07  & 0.07
		& 0.04  & 0.07  & 0.09  & 0.12  & 0.15 \\
		CCA+SpClust(GC)  
		& 0.15  & 0.13  & 0.12  & 0.11  & 0.09  
		& 0.09  & 0.13  & 0.16  & 0.20  & 0.21 \\
		3VCCA+SpClust(GC) 
		& 0.16  & 0.14  & 0.13  & 0.11  & 0.09
		& 0.09  & 0.14  & 0.17  & 0.20  & 0.23 \\
		MMC(GC)    
		& 0.10  & 0.09  & 0.09  & 0.07  & 0.07 
		& 0.06  & 0.07  & 0.11  & 0.12  & 0.17 \\	
		DCCA(GC)
		& 0.18  & 0.12  & 0.11  & 0.09  & 0.09
		& 0.10  & 0.13  & 0.15  & 0.18  & 0.19 \\
		DCCAE(GC)
		& 0.18  & 0.13  & 0.11  & 0.09  & 0.09
		& 0.10  & 0.13  & 0.15  & 0.18  & 0.19 \\
		\hline
		HML-RF(GC)
		& 0.20  & 0.15  & 0.13  & 0.10  & 0.09 
		& 0.11  & 0.14  & 0.16  & 0.18  & 0.19 \\	
		HML-RF(LN) 
		& 0.29  & 0.20  & 0.17  & 0.14  & 0.11
		& 0.15  & 0.20  & 0.23  & 0.25  & 0.26 \\
		HML-RF(AM)
		& \textbf{0.34} & \textbf{0.24} & \textbf{0.20} & \textbf{0.17} & \textbf{0.15} 
		& \textbf{0.18} & \textbf{0.24} & \textbf{0.28} & \textbf{0.30} & \textbf{0.32} \\
		
		\hline
	\end{tabular}
\end{table*}

%

\subsection{Evaluation on Completing Local Instance-Level Concept Structure}
\label{sec:evaluate_completion}

\noindent 
\textbf{Baseline methods}:
We compared our missing tag completion method (all three algorithms)
for completing local instance-level semantic concept 
against the following three contemporary approaches:
(1) Linear Sparse Reconstructions (LSR) \cite{lin2013image}: 
A state-of-the-art image-specific and tag-specific Linear Sparse Reconstruction scheme for tag completion.
(2) Tag Completion by Matrix Recovery (TCMR) \cite{feng2014image}: 
A recent tag matrix recovery based completion algorithm 
that captures both underlying tag dependency and visual consistency.
(3) A group of cluster based completion methods: 
Specifically, we used the same algorithm as HML-RF(GC) for 
missing tag recovery (Section \ref{sec:method_tag_completion}).
The clusters were obtained by the compared methods in Section \ref{sec:clustering}.
\zhu{For HML-RF, we utilised the clustering results by the five-layer hierarchy.}
Similarly, we name these completion methods in form of ``ClusteringMethodName(GC)'', e.g. MMC(GC).

\vspace{0.1cm}
\noindent
\textbf{Evaluation metrics}:
We utilised three performance measures:
(1)	{AP@$N$},
which measures Average Precision of $N$ recovered tags. 
(2)	{AR@$N$},
which calculates Average Recall of $N$ recovered tags, i.e. the percentage of correctly recovered tags over all ground truth missing tags.
(3)	{Coverage@$N$},
which denotes the percentage of samples with at least one correctly recovered tag
when $N$ tags are completed.

\begin{figure*} 
	\centering
	\includegraphics[width=0.99\linewidth]{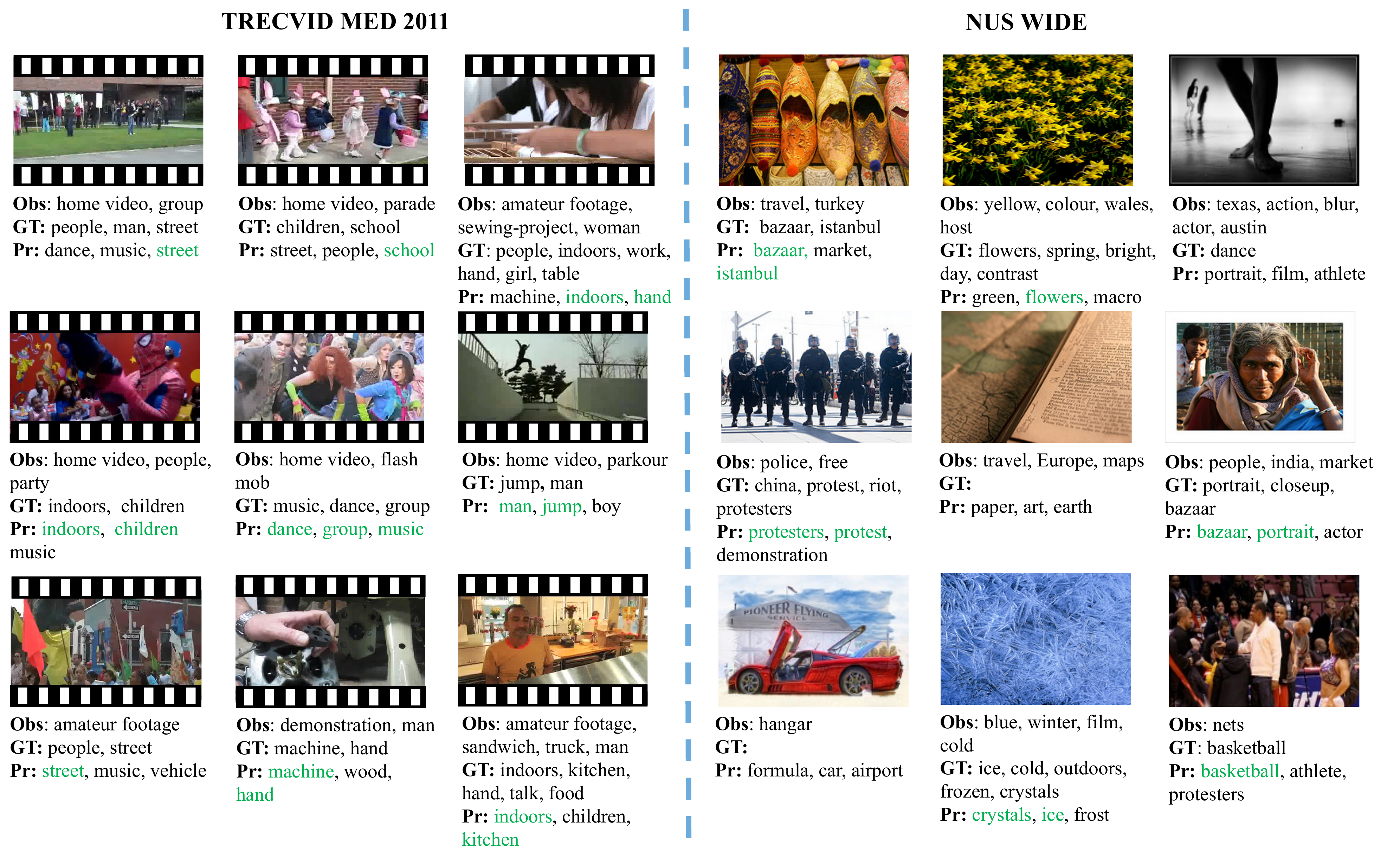}
	\vskip -0.3cm
	\caption{
		\zhu{
			Examples of tag completion by our HML-RF(AM) method.
			Correctly recovered tags are highlighted in green colour.
			(Obs: Observed tags; GT: Ground Truth for missing tags; Pr: Predicted tags).
		}
	}
	\label{fig:completion_example}
\end{figure*}

\subsubsection{Missing Tag Completion Evaluation on TRECVID}
\label{sec:recover_tag_TRECVID}
The tag completion results on TRECVID MED 2011 are given in Tables \ref{tab:AP_AR_TRECVID} and \ref{tab:coverage}.
It is evident that the proposed completion algorithms outperform all compared methods.
\zhu{In particular, it is observed that global clusters provide strong cues for missing tag recovery, 
e.g. DCCAE is superior than or similar to the state-of-the-art completion methods 
TCMR and LSR at AP@$1$.}
This suggests the intrinsic connection between global and
local semantic structures, and validates our motivation for bridging
the two visual data structure analysis tasks (Section \ref{sec:method_tag_completion}). 
By more accurate global group structure revelation,
HML-RF(GC) enables even better missing tag completion, e.g.
obtaining higher average precision and recall than other clustering methods.
Moreover, HML-RF(GC) produces better tag recovery than our local neighbourhood based completion method HML-RF(LN), particularly in cases of completing multiple tags. 
This further indicates the positive restricting effect of global data structures 
over inferring local instance-level semantic concept structures.
However, HML-RF(LN) provides best AR@$1$, which 
should be due to its strict rule on selecting neighbourhoods. 
While TCMR considers both tag correlation as well as visual consistency,
it is still inferior to the proposed HML-RF owing potentially to 
(1) the incapability of exploiting the tag
    abstract-to-specific hierarchy knowledge; and 
(2) the assumptions on low rank matrix recovery may be not
    fully satisfied given real-world visual data. 
These observations and analysis demonstrate the superiority of
    our HML-RF in instance-level tag completion, owing to its 
    favourable capability in jointly
    learning heterogeneous visual and tag data 
    and thus more accurate semantic visual 
    structure disclosure.

\subsubsection{Missing Tag Completion Evaluation on NUS-WIDE}
\label{sec:recover_tag_NUS}

Tables \ref{tab:AP_AR_NUS} and \ref{tab:coverage} show the comparative
results for tag completion on the NUS-WIDE image dataset
\cite{nus-wide-civr09}, where the available tags are more sparse
($0.48\%$) as compared to the TRECVID MED 2011 video dataset ($3.5\%$).  
\zhu{Overall, our methods HML-RF(AM) outperforms all other baselines,
including the state-of-the-art models LSR and TCMR, 
and contemporary deep-based
multi-modal correlation learning methods DCCA and DCCAE}. 
We found that
our HML-RF(GC) model dose not perform as strongly as on TRECVID MED 2011.
This shall be due to less accurate global group structures discovered (see Table \ref{tab:cluster_nus}).
By imposing stringent neighbourhood selection, 
HML-RF(LN) produces considerably better tag recovery accuracy than HML-RF(GC).
This validates the proposed pure neighbourhood based completion strategy
in handling sparse and incomplete tags where a large number of missing tags
can negatively bias tag recovery (Section \ref{sec:method_tag_completion}).
HML-RF(AM) achieves the best results 
due to the combined benefits from both local and global neighbourhood structures.
These evaluations and observations further validate the capability and
efficacy of the proposed model in jointly learning heterogeneous
visual and tag modalities and semantically interpreting the instance-level concept
structure of ambiguous visual content in both video and image data.
For qualitative evaluation, we show in Figure \ref{fig:completion_example}
the top-$3$ recovered tags per sample by our HML-RF(AM) method.

\section{Conclusion}
In this work, we presented an visual concept structure discovery framework
by formulating a novel Hierarchical-Multi-Label Random Forest (HML-RF) model 
for jointly exploiting heterogeneous visual and tag data modalities,
with the aim of creating an intelligent visual machine for automatically 
organising and managing large scale visual databases.
The proposed new forest model, 
which is defined by a new information gain function,
enables naturally incorporating tag abstractness hierarchy and 
effectively exploiting multiple tag statistical correlations,
beyond modelling the intrinsic interactions between visual and tag modalities.
With the learned HML-RF, we further derive a generic clustering pipeline for global group structure discovery and three tag completion algorithms for local instance-level tag concept structure recovery.
%
%
Extensive comparative evaluations 
have demonstrated the advantages and superiority of the proposed approach 
over a wide range of
existing state-of-the-arts clustering, multi-view embedding and tag completion models, 
particularly in cases where only sparse tags are accessible. 
Further, a detailed model component examination is provided 
for casting insights on our modelling principles and model robustness. 
In addition to the above two applications, our HML-RF model can potentially benefit
other related problems, such as retrieval and manifold ranking.

\section*{Acknowledgements}
This work was partially supported by the China Scholarship Council, Vision Semantics Limited, and Royal Society Newton Advanced Fellowship Programme (NA150459).
The corresponding author is Xiatian Zhu.

\section*{References}

\bibliography{ref}

\end{document}